\documentclass[10pt,journal,compsoc]{IEEEtran}

\ifCLASSOPTIONcompsoc
  \usepackage[nocompress]{cite}
\else
  \usepackage{cite}
\fi

\ifCLASSINFOpdf
\else
\fi

\usepackage[utf8]{inputenc}
\usepackage{amsmath}
\usepackage{graphicx}
\usepackage{xspace}
\usepackage{multirow}
\usepackage{multicol}
\usepackage{booktabs}
\usepackage{xcolor}
\usepackage{colortbl}
\usepackage{amssymb}

\title{Adversarial Contrastive Learning for Evidence-aware Fake News Detection with Graph Neural Networks}

\author{Junfei Wu, 
        Weizhi Xu, 
        Qiang Liu, \IEEEmembership{Member, IEEE}, \\
        Shu Wu, \IEEEmembership{Senior Member, IEEE},
        and Liang Wang, \IEEEmembership{Fellow, IEEE}
\IEEEcompsocitemizethanks{
\IEEEcompsocthanksitem This paper is an extended version of GET \cite{xu2022evidenceaware}, which has been published in the Proceedings of the ACM Web Conference 2022.
\IEEEcompsocthanksitem All authors are with the Center for Research on Intelligent Perception and Computing, National Laboratory of Pattern Recognition, Institute of Automation, Chinese Academy of Sciences and are also with the School of Artificial Intelligence, University of Chinese Academy of Sciences. 
\IEEEcompsocthanksitem The first two authors contributed equally to this work
\IEEEcompsocthanksitem Corresponding author: 
Qiang Liu (qiang.liu@nlpr.ia.ac.cn).
}% <-this % stops a space
\thanks{Manuscript received April 19, 2005; revised August 26, 2015.}}

\newcommand{\modelname}{\textbf{\underline{G}}raph-based s\textbf{\underline{E}}mantic structure mining framework with Con\textbf{\underline{TRA}}stive \textbf{\underline{L}}earning}
\newcommand{\themodel}{GETRAL\xspace}

\begin{document}

\IEEEtitleabstractindextext{%
\begin{abstract}
The prevalence and perniciousness of fake news have been a critical issue on the Internet, which stimulates the development of automatic fake news detection in turn. In this paper, we focus on the evidence-based fake news detection, where several evidences are utilized to probe the veracity of news (i.e., a claim). Most previous methods first employ sequential models to embed the semantic information and then capture the claim-evidence interaction based on different attention mechanisms. Despite their effectiveness, they still suffer from three weaknesses. Firstly, due to the inherent drawbacks of sequential models, they fail to integrate the relevant information that is scattered far apart in evidences for veracity checking. Secondly, they underestimate much redundant information contained in evidences that may be useless or even harmful. Thirdly, insufficient data utilization limits the separability and reliability of representations captured by the model, which are sensitive to local evidence.
To solve these problems, we propose a unified \modelname, namely \themodel in short. Specifically, different from the existing work that treats claims and evidences as sequences, we first model them as graph-structured data and capture the long-distance semantic dependency among dispersed relevant snippets via neighborhood propagation. After obtaining contextual semantic information, our model reduces information redundancy by performing graph structure learning. Then the fine-grained semantic representations are fed into the downstream claim-evidence interaction module for predictions. 
Finally, the supervised contrastive learning accompanied with adversarial augmented instances is applied to make full use of data and strengthen the representation learning. 
% by enlarging the gap between different claim-aware interaction relations.
Comprehensive experiments have demonstrated the superiority of \themodel over the state-of-the-arts and validated the efficacy of semantic mining with graph structure and contrastive learning.
\end{abstract}

}

\maketitle

% \ifCLASSOPTIONcompsoc
% \IEEEraisesectionheading{\section{Introduction}\label{sec:introduction}}
% \else
% \section{Introduction}
% \label{sec:introduction}
% \fi

\section{Introduction}

% for IEEE Computer Society journal papers produced under \LaTeX\ using
% IEEEtran.cls version 1.8b and later.
% % You must have at least 2 lines in the paragraph with the drop letter
% % (should never be an issue)
% I wish you the best of success.\cite{Allcott2017SocialMA}

 

% \subsection{subsection}

\IEEEPARstart{S}{ocial} media has facilitated the dissemination and exchange of information, thus profoundly reshaping the convention of people to consume information. However, due to the inability to verify lots of real-time information, social media has also become a hotbed of fake news, which is always fabricated by making some minor changes to the correct statement. Fake news is not only highly deceptive but also inflammatory, potentially influencing real-world events. The widespread of fake news in diverse domains, such as politics \cite{Allcott2017SocialMA} and public health \cite{Naeem2020TheC}, has posed a huge threat to web security and human society. Therefore, the research on automatic fake news detection is challenging and in demand.

Generally, previous methods could be roughly categorized into two groups, i.e., pattern-based approaches and evidence-based approaches \cite{Sheng2021IntegratingPA}. The former methods regard the fake news detection as a feature recognition task, where language models are employed to verify the veracity of news solely according to the text pattern, e.g., writing styles. 
% For example, a large proportion of fake news contains exaggerated diction. 
However, pattern-based methods usually suffer from the poor generalization and interpretability. The latter approaches model the task as a reasoning process, where external evidences are provided to probe the veracity of a claim. Models are required to discover and integrate useful information in given evidences for claim verification.

\begin{figure}[t]
  \begin{center}
  \includegraphics[width=0.45\textwidth]{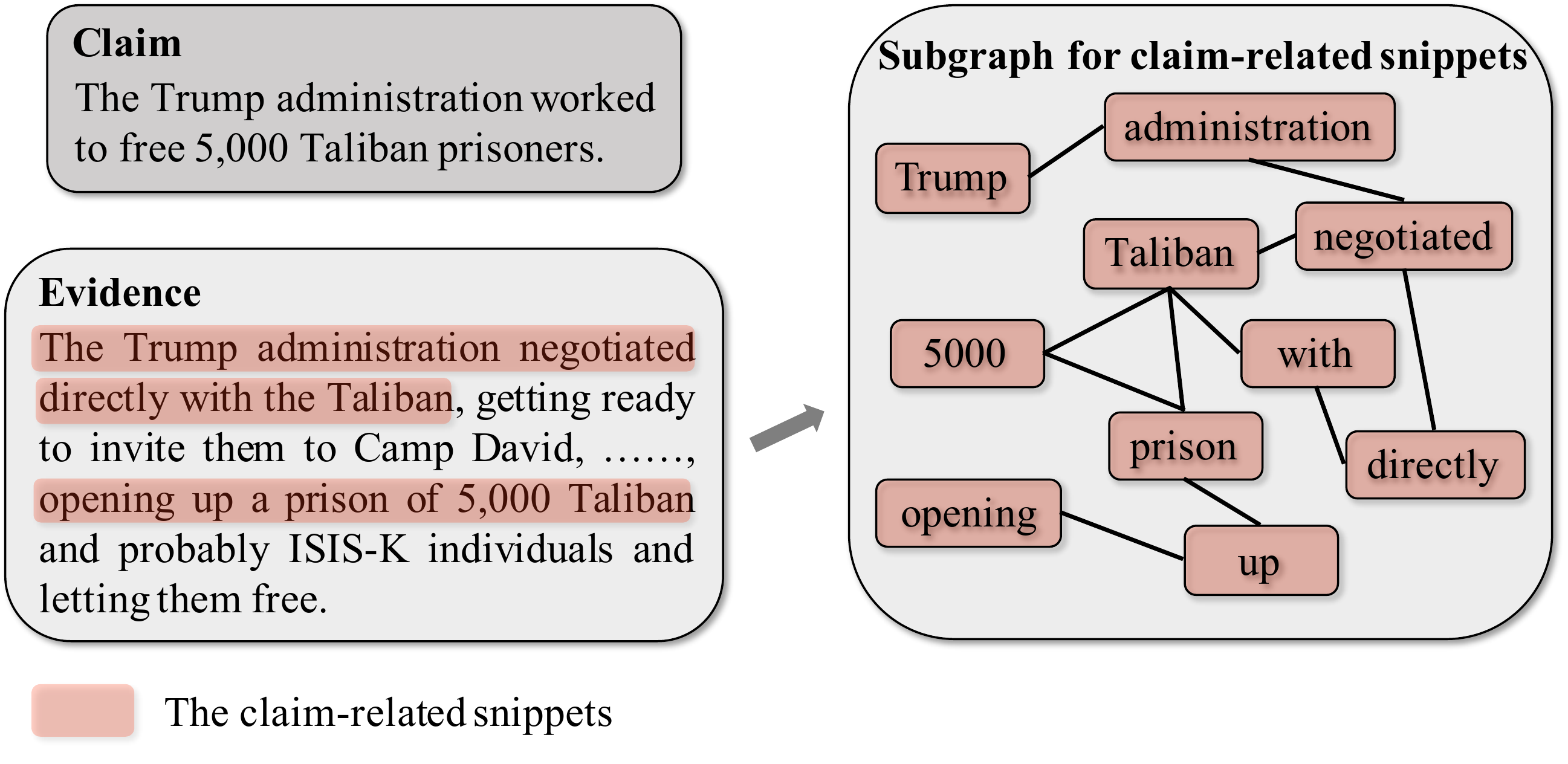}
  \end{center}
   \caption{A toy example where a claim and its relevant evidence are given. Two significant snippets for verifying the claim are highlighted (``.....'' represents that we omit several sentences for conciseness). The right graph is constructed according to the highlighted snippets. Such two snippets have a long distance in the plain text while they are pulled close on the constructed semantic graph via the shared keyword ``Taliban''. Besides, there is much redundant information (texts except the highlighted parts), which is useless for claim verification.}
   \label{fig:example}
\end{figure}

In this paper, we focus on the evidence-based pipeline. Existing methods usually follow a two-step paradigm: 1) they first capture the semantics of claims and evidences separately. 2) Next, they model the claim-evidence interaction to explore the semantic coherence or conflict for more accurate and interpretable verdict. To name a few representative models, the pioneering work DeClarE \cite{popat2018declare} utilizes bidirectional LSTMs to model textual features, followed by a word-level attention mechanism to capture the claim-evidence interaction. HAN \cite{ma2019sentence} further considers the sentence-level interaction to explore more general semantic coherence. To obtain multi-level semantic interaction, some recent works \cite{vo2021hierarchical, wu2021unified} employ hierarchical attention networks. 

Nevertheless, existing work focuses on the specific design of different interaction models (the second step) while neglecting exploring fine-grained semantics of claims and evidences (the first step). 
In addition, they ignore sufficient data utilization for capturing separable and reliable representations. To be specific, we argue that there are three main weaknesses in previous methods,

(1) The complex, long-distance semantic dependency is less explored. Taking Figure \ref{fig:example} as an example, two highlighted snippets are separated by plenty of words, which induces a long distance between them. Such snippets both contain important information for verifying the claim, i.e., the subject ``The Trump administration'' and the action ``opening up a prison of 5,000 Taliban''. Therefore, fusing the information is indispensable and beneficial for claim veracity prediction. However, the long-distance semantic dependency between such information is hard to be captured due to the inherent drawbacks of sequential models utilized in previous methods. 

(2) Existing methods pay little attention to the redundant information involved in semantics. Such redundancy is useless or even harmful for fake news detection, e.g., as depicted in Figure \ref{fig:example}, a large number of text segments, such as ``getting ready to invite them to Camp David'', have no substantial contribution to the news veracity checking. Though previous models employ attention mechanisms to reduce the effect of unrelated words, these irrelevant texts are still preserved, which may introduce noises to the downstream claim-evidence interaction, deteriorating the final performance of veracity checking. 
An intuitive solution is to discard words with low attentive scores based on previous methods. However, 
% they compute the score for each word independently, ignoring the complex semantic structure among words.  
we argue that it is significant to model the redundancy with rich semantic structural information, as the redundancy is not only related to the self-information, but also induced by its contexts.

(3) Previous works mainly focus on the learning of mapping from the interaction representations to veracity labels, without sufficient data utilization. This is likely to limit the separability of representations learned in the model. Moreover, insufficient data utilization also causes the sensitivity of the high-level representations to slight changes in the retrieved evidence. In real-world scenes, as the evidence retrieved online may contain much noise and change under different conditions, the captured representations are not reliable and may degrade the detection performance. In addition, the attention mechanism is always utilized for selecting the most useful evidence, which aggravates the sensitivity of the captured representations to some significant evidence. Therefore, we argue that taking full use of data for efficient training can obtain more separable and reliable representations, which contain core clues instead of being sensitive to local evidence.

% However, existing methods only focus on the learning of mapping from interaction representations to labels, while ignoring the efficient learning and utilization of the data. Therefore, we should make full use of these relationships between different samples to learn more separable representations. In the meantime, emphasizing the fitting of labels also makes the model too sensitive to slight changes in evidence, which is often affected by the evidence retrieval process. Hence, we should also strengthen the use of data itself to avoid the model being too sensitive.

To tackle the aforementioned problems, we propose a unified \modelname, namely \themodel for exploring fine-grained semantics and capturing enhanced representations.  Firstly, modeling sequential data as graphs has benefited many tasks, such as text classification \cite{Yao2019GraphCN, texting} and sequential recommendation \cite{srgnn}, owing to its capability of capturing long-distance structural dependency. To this end, we firstly utilize graph structure to model both claims and evidences, where nodes indicate words and edges represent the co-occurence between two words. Thereafter, the dispersed claim-related snippets are pulled close on graphs, thus the useful information could be better fused via neighborhood propagation. For example, in Figure \ref{fig:example}, after constructing the graph for two highlighted snippets distant from each other in plain texts, they are pulled close via the shared keyword ``Taliban'' so that the long-distance semantic dependency can be captured. 

Moreover, to alleviate the negative impact of redundant information, within our graph-based framework, we treat the redundancy mitigation as a graph structure learning process, where unimportant nodes are discarded according to complex semantic structures including both self-features and their contexts. The former is related to its own information and its relevance to the claim, and the latter is related to its graph topology. To date, our graph-based framework has captured the fine-grained semantics via long-distance dependency modeling and redundancy mitigation. 
% Based on such semantics, we can apply the widely used attention mechanism in previous work to readout node features and form the claim- and evidence-level representations, followed by claim-evidence interactions to integrate information for the final veracity prediction. 

Finally, inspired by the success of recent work \cite{chen2021cil, zhang2021supporting} which integrates contrastive learning, we introduce a supervised contrastive learning auxiliary task to strengthen the representation learning. In addition, to reduce the sensitivity to local evidence, we employ adversarial gradient perturbation \cite{miyato2016adversarial, li2022hiclre} to augment the contrastive instances at the feature level. In specific, with the veracity label utilized as the supervised signal, the representations of claim-evidence interactions of the same class are pulled close, while those of different classes are pulled apart. Subsequently, the reliable representations to discriminate different claim-evidence interactions can be better captured. 

% To summarize, in this article, we first review several related works in Section \ref{related_work}. Then we formulate the task of evidence-aware fake news detection in Section \ref{task} and elaborate each component of \themodel in Section \ref{model}. We also perform extensive experiments to verify the effectiveness of our method and analyze the results in Section \ref{experiments}. 

Our main contributions can be summarized as follows:
\begin{itemize}
    \item We model claims and evidences as graph-structured data and introduce a simple and effective graph structure learning approach for redundancy mitigation. The captured long-distance and fine-grained semantics based on the structure can boost the performance of downstream interaction models.
    \item We introduce a supervised contrastive learning task and integrate the adversarial gradient perturbation for efficient training. Then the captured representations are more separable and reliable for detection.
    \item Comprehensive experiments are conducted to demonstrate the superiority of \themodel and the effectiveness of each component. Our code is available at https://github.com/CRIPAC-DIG/GETRAL
\end{itemize}

\section{related work}
\label{related_work}
% In this section, we briefly review previous work in three related domains: graph neural networks, fake news detection and contrastive learning.

\subsection{Graph Neural Networks}
Graph neural networks (GNNs) learn the node representation by gathering information from the neighborhood, i.e., neighborhood propagation/aggregation. Current GNNs can be roughly divided into two groups, namely spectral approaches \cite{Defferrard2016ConvolutionalNN, Kipf2017SemiSupervisedCW} and spatial approaches \cite{Velickovic2018GraphAN, Hamilton2017InductiveRL}. Owing to the capability of capturing long-distance structural relationship on graphs, GNNs have been widely utilized and achieved satisfactory performance in several tasks, such as recommender system \cite{srgnn, Chen2020HandlingIL, zhang2020personalized, Zhang2021MiningLS}, text classification \cite{Yao2019GraphCN, texting}, and sentiment analysis \cite{Wang2020RelationalGA, Li2021DualGC}. 

Recently, researchers have observed that graphs inevitably contain noises that may deteriorate the training of GNNs \cite{Jin2020GraphSL}. To handle this problem, graph structure learning (GSL) is proposed, aiming to jointly learn an optimized graph structure and node embeddings. Existing GSL methods mainly fall into three groups \cite{zhu2021gsl}: 1) \emph{the metric-learning-based methods} where the adjacency matrices are built as metrics coupled with node embeddings. Therefore, the graph topology is updated with node embeddings being optimized. The metrics are mainly defined as the attention-based function \cite{jiang2019gsl, chen2020gsl, cosmo2020gsl} or kernel function \cite{li2018gsl, wu2018gsl}. 2) \emph{the probabilistic methods} assume that the adjacency matrix is generated by sampling from a specific probabilistic distribution \cite{franceschi2018gsl, franceschi2019gsl, Zhang2019gsl}. 3) \emph{the direct-optimized methods} treat the graph topology as learnable parameters that are updated together with task-specific parameters simultaneously, without depending on preset priors (namely node embeddings and distributions in the first two groups, respectively). The topology is optimized with the guidance of task-specific objectives (and some normalization constraints) \cite{Yang2019gsl, Jin2020GraphSL}. It is worth noting that existing graph pooling methods \cite{ying2018hierarchical, gao2019graph, lee2019self} could also be viewed as GSL algorithms, since the pooling target is to keep the most valuable nodes that preserve the graph structural information well, where the graph structure is optimized via merging or dropping nodes. Besides, GNNs are widely employed in the domain of fact verification, which have achieved promising performance \cite{Zhou2019GEARGE, Liu2020FinegrainedFV, Zhong2020ReasoningOS}. Though fact verification is similar to fake news detection on the task setting, the latter requires more fine-grained semantics since the texts consist of more redundancy.

\subsection{Fake News Detection}
Several fake news detection methods have been proposed recently, which can be roughly grouped into two categories.

The first is the pattern-based pipeline where models solely consider the text pattern involved in the news itself. Different works always focus on different kinds of patterns. Popat et al. \cite{popat2016} classify a claim as true or fake in accordance with stylistic features and the article stance. Besides, some researchers attempt to verify the truthiness via the feedback in social media, such as reposts, likes, and comments \cite{Yu2017ACA, volkova-etal-2017-separating, Vo2018, Benamira2019SemiSupervisedLA, Chandra2020GraphbasedMO, Jin2021TowardsFR, liu2018mining}. Recently, more attention has been paid to the emotional pattern mining, where it holds an assumption that there are probably obvious sentiment biases in fake news \cite{Ajao2019, Gia2019, zhang2021www, zhu2022memory}.

The second is the evidence-based pipeline where researchers propose to explore the semantic similarity (conflict) in claim-evidence pairs to check the news veracity. Evidences are usually retrieved from the knowledge graph \cite{vlachos-riedel-2015-identification} or fact-checking websites \cite{vlachos-riedel-2014-fact} by giving unverified claims as queries. DeClarE \cite{popat2018declare} is the first work to utilize evidences in fake news detection. It employs BiLSTMs to embed the semantics of evidences and obtains the claim's sentence-level representation via average pooling. Next, it introduces an attention-based interaction to compute the claim-aware score for each word in evidences. Similar to the pioneering work, the following methods utilize the sequential models to obtain the semantic embeddings, followed by attention mechanisms performed on different granularities. HAN \cite{ma2019sentence} computes the sentence-level coherence and entailment scores between claims and evidences. EHIAN \cite{wu2020evidence} employs the self-attention mechanism to obtain word-level interaction scores. Recent works \cite{vo2021hierarchical, wu2021unified, Wu_Rao_Sun_He_2021} hierarchically integrate both word-level and sentence-level interactions into the final representation for verification. In summary, they all employ sequential models to embed semantics and apply attention mechanisms to capture the claim-evidence interactions.

Different from existing works, we propose a unified graph-based model, where the long-distance semantic dependency is captured via constructed graph structures and the redundancy is reduced by graph structure learning. 

\subsection{Contrastive Learning}
% 1. definition, 2. current state & application, 3. our work
Contrastive learning is an effective training paradigm that captures separable and distinguishable representations which can bring significant improvement for downstream tasks. Specifically, it utilizes InfoNCE loss \cite{oord2018representation} to pull the representations of positive samples closer while pulling the negative samples apart, forming a representation space with alignment and uniformity. Nowadays, contrastive learning has been applied to several tasks \cite{oord2018representation, chen2020simple, he2020momentum, khosla2020supervised}.

Contrastive learning was first introduced for training visual representations \cite{bachman2019learning, he2020momentum, chen2020simple}, by conducting visual augmentations including cropping, resizing and other operations at the visual level to obtain positive pairs, while different instances naturally form the negative pairs with each other. Subsequently, supervised contrastive learning \cite{khosla2020supervised} has also gained much attention. It integrates the class relationship to construct contrastive instances to calibrate representations. For graph data, perturbations are imposed on the graph topology and node 
features to generate corrupted views as the contrastive instances. Then, the graph representations which better contain structure and semantic information are obtained by maximizing the agreement between either global graph embeddings or local node embeddings \cite{velickovic2019deep, you2020graph, zhu2020deep, zhu2021graph}. In addition, as it is critical for natural language processing to compress dense semantics, contrastive learning has been introduced to leverage abundant textual resources to learn better representations. SimCSE  \cite{gao2021simcse} uses only dropout as minimal data augmentation to sentence embeddings and boost the performance of pretrained model significantly. In distantly supervised relation extraction, CIL \cite{chen2021cil} and HiCLRE \cite{li2022hiclre} exploit the abundant instance relations and propose contrastive instance learning to obtain accurate representations under noise efficiently. Contrastive learning has also been adopted by pattern-based fake news detection for domain adaptation \cite{lin2022detect, yue2022contrastive}. Different from these works, we combine a supervised contrastive learning task with the classification task to capture more separable representations for evidence-aware fake news detection.

\section{method}
\label{method}
% In this section, we first give the task formulation. Then, we introduce the proposed model \themodel in detail, as depicted in Figure \ref{fig:model}.

\begin{figure*}[t]
   \begin{center}
%   \vspace{-4mm}
   \includegraphics[width=0.90\textwidth]{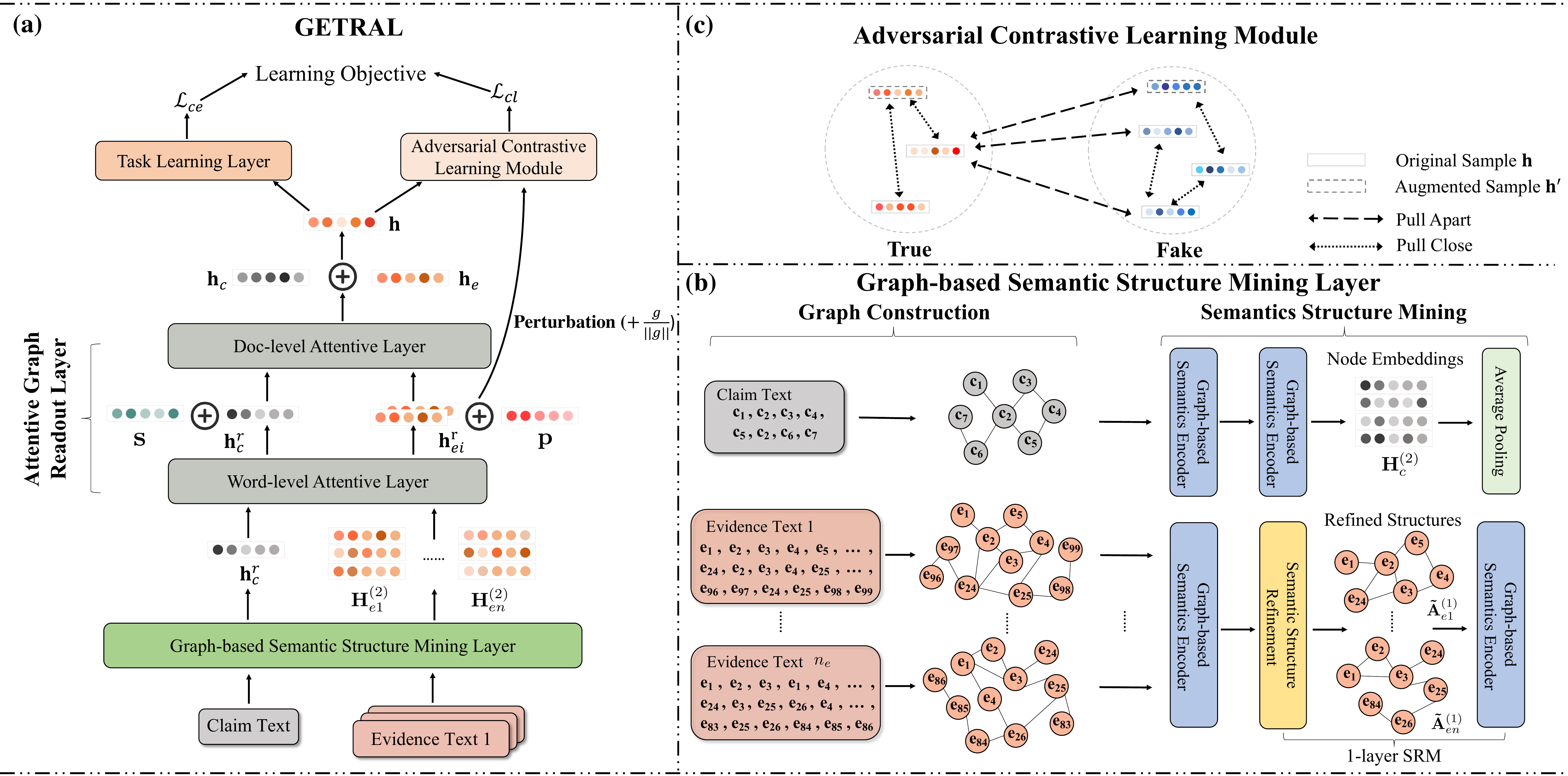}
   \end{center}
   %\vspace{-1.0cm}
   \caption{
   (a) The overall architecture of \themodel. It consists of Graph-based Semantic Structure Mining Layer, Attentive Graph Readout Layer, Task Learning Layer and Adversarial Contrastive Learning Module.
   %The architecture of \themodel. Firstly, the plain texts are input into Graph-based Semantic Structure Mining Layer to capture fine-grained semantics. Then claim and evidence embeddings along with their speaker and publisher information are fed into the multi-level attentive layers to obtain the final representations. These representations are further enhanced by contrastive learning module.
   (b) The Graph-based Semantic Structure Mining Layer first transform plain texts into graphs, then perform neighbourhood aggregation and structure learning to obtain fine-grained semantics.
   %(the window size is 2 in the figure).The same words repeatedly appear in texts are merged into one node. Next, we introduce graph-based semantics encoder to capture long-distance structural dependencies and generate high-order representations. Furthermore, the semantic structure refinement layer is proposed to generate optimized structures \(\{\mathbf{\tilde{A}}^{(1)}_{e1}, \ldots, \mathbf{\tilde{A}}^{(1)}_{en}\}\) for \(n\) evidences, where redundant nodes are discarded. 
%   (The 1-layer SRM consists of a semantic structure refinement layer and a graph-based semantics encoder). Thereafter, the fine-grained semantics is obtained on refined graphs. 
    (c)  The Adversarial Contrastive Learning Module pulls the representations of positive pairs close while pushing representations of negative pairs apart.
   }
   \label{fig:model}
\end{figure*}

\subsection{Task Formulation}
\label{task}
Evidence-based fake news detection is a classification task, where the model is required to output the prediction of news veracity. Specifically, the inputs are a claim \(c\), several related evidences \(\mathcal{E} = \left\{e_1, e_2, \ldots, e_{n}\right\}\), and their corresponding speakers \(\mathbf{s} \in \mathbb{R}^{1 \times b}\) or publishers \(\mathbf{p} \in \mathbb{R}^{n \times b} \),
where \(n\) is the number of evidences and \(b\) is the dimension of speaker and publisher embeddings. The output is the predicted probability of veracity \(\hat{y} = f\left(c, \mathcal{E}, \mathbf{s}, \mathbf{p}, \Theta \right)\), where \(f\) is the verification model and \(\Theta\) is its trainable parameters.
% \(c = \left\{c_1, c_2, \ldots, c_{l_c}\right\}\)

\subsection{The Proposed Model: \themodel}
\label{model}
In this part, we elaborate our unified graph-based model \themodel, which can be mainly separated into five modules: 1) \emph{Graph Construction}, 2) \emph{Graph-based Semantics Encoder}, 3) \emph{Semantic Structure Refinement}, 4) \emph{Attentive Graph Readout Layer}, and 5) \emph{Adversarial Contrastive Learning Module}. 
% The overall structure of \themodel is shown in Figure \ref{fig:model}.
\subsubsection{Graph Construction}
In order to capture the long-distance dependency of relevant information, we first convert the original claims and evidences to graphs. Like previous graph-based methods in other NLP tasks \cite{Yao2019GraphCN, grmm, ghrm, texting}, we use a fix-sized sliding window to screen out the connectivity for each word on graphs. In detail, the center words in every window will be connected with the rest of words in it (if connected, the corresponding entry in the adjacency matrix is 1, otherwise 0), which captures the local context in the center word's neighborhood. Furthermore, to model the long-distance dependency, we merge all the same words into one node on graph, which explicitly gathers their local contexts (e.g., the word \(e_2\) in evidence text 1 in Figure \ref{fig:model}). Therefore, several relevant snippets that scatter far apart is close on graphs, which can be explored via the high-order message propagation. In addition, the initial node representations are the corresponding word embeddings. Note that we also try to construct a graph in a fully-connected or semantic-similarity-based manner, but these two ways are inferior to the sliding-window-based method, which may be due to the redundant noises induced by the dense connection.

To ensure the numerical stability, we perform Laplacian normalization on adjacency matrices, denoted as \(\tilde{\mathbf{A}} = \mathbf{D}^{-\frac{1}{2}} (\mathbf{A}+\mathbf{I}) \mathbf{D}^{-\frac{1}{2}}\), where \(\mathbf{D}\) is the diagonal degree matrix (i.e., \(\mathbf{D}_{ii} = \sum_j \mathbf{A}_{ij}\)) and \(\mathbf{I}\) is the identical matrix. Finally, we denote the initial normalized adjacency matrices and node feature matrices of claim and evidence as \(\tilde{\mathbf{A}}^{(0)}_c \in \mathbb{R}^{N_c \times N_c}\), \(\tilde{\mathbf{A}}^{(0)}_e \in \mathbb{R}^{N_e \times N_e}\) and \(\mathbf{H}^{(0)}_c \in \mathbb{R}^{N_c \times d}\), \(\mathbf{H}^{(0)}_e \in \mathbb{R}^{N_e \times d}\), respectively. \(N_c\) and \(N_e\) is the number of nodes in initial claim and evidence graphs, \(d\) is the dimension of word embeddings.

Taking the established graph structures and node embeddings as inputs, we design a graph-based model to better capture complex semantics and obtain refined semantic structures.

\subsubsection{Graph-based Semantics Encoder}
To mine the long-distance semantic dependency, we propose to utilize GNNs as the semantics encoder.
In particular, as we expect to adaptively keep a balance between self-features and the information of neighboring nodes, we employ graph gated neural networks (GGNN) to perform neighborhood propagation on both claim and evidence graphs, enabling nodes to capture their contextual information, which is significant for learning high-level semantics. Formally, it can be written as follows:
\begin{align}
    \label{eq:ggnn-s}
    \mathbf{a}_{i}&=\sum_{(w_{i}, w_{j}) \in \mathcal{C}} \mathbf{\tilde{A}}_{ij} \mathbf{W}_{a} \mathbf{H}_{j} \\
    \mathbf{z}_{i}&=\sigma\left(\mathbf{W}_{z} \mathbf{a}_{i}+\mathbf{U}_{z} \mathbf{H}_{i}+\mathbf{b}_{z}\right) \\
    \mathbf{r}_{i}&=\sigma\left(\mathbf{W}_{r} \mathbf{a}_{i}+\mathbf{U}_{r} \mathbf{H}_{i}+\mathbf{b}_{r}\right) \\
    \tilde{\mathbf{H}}_{i}&=\tanh \left(\mathbf{W}_{h} \mathbf{a}_{i}+\mathbf{U}_{h}\left(\mathbf{r}_{i} \odot \mathbf{H}_{i}\right)+\mathbf{b}_{h}\right) \\
    \mathbf{\hat{H}}_{i}&=\tilde{\mathbf{H}}_{i} \odot \mathbf{z}_{i}+\mathbf{H}_{i} \odot\left(1-\mathbf{z}_{i}\right) 
    \label{eq:ggnn-e}
\end{align}
where \(\mathcal{C}\) denotes the edge set, \(\mathbf{W}_*\), \(\mathbf{U}_*\), and \(\mathbf{b}_*\) are trainable parameters, which control the proportion of the neighborhood information and self-information. \(\sigma\) is the non-linear activation unit and we utilize the Sigmoid function in our model. For brevity, we denote Eq. (\ref{eq:ggnn-s}) - (\ref{eq:ggnn-e}) as \(\textbf{GGNN}(\mathbf{\tilde{A}}, \mathbf{H})\)\footnote{When generally describing the module that will be repeatedly utilized in the model, we omit the superscripts indicating layer number for brevity.}.

\subsubsection{Semantic Structure Refinement}
As evidences always contain redundant information that may mislead the model to focus on unimportant features, it is beneficial to discover and filter out the redundancy, thus obtaining refined semantic structures. To this end, in our graph-based framework, we treat the redundancy mitigation as a graph structure learning process, whose aim is to learn the optimized graph topology along with better node representations. Previous GSL methods generally optimize the topology in three ways, i.e., dropping nodes, dropping edges, and adjusting edge weights. Since the redundancy information is mainly involved in words denoted as nodes in evidence graphs, we attempt to refine evidence graph structures via discarding redundant nodes, inspired by previous GSL methods \cite{lee2019self, chen2020gsl, Zhang2019gsl}. 

In particular, we propose to compute a redundancy score for each node, based on which we obtain a ranking list and the nodes with the top-\(k\) redundancy scores will be discarded, then we adjust the aggregation weight for the rest nodes. For each node, its redundancy can be determined by its own information and its relevance to the claim. Hence, we evaluate the independent information redundancy of each node from the view of the node itself and claim relevance, respectively. Specifically, the node-self redundancy is directly measured by a linear projection. To obtain the claim-related redundancy, we utilize the Gaussian kernel to measure the fine-grained relevance of each token in the evidence to all tokens in claims. It has been proved in \cite{Liu2020FinegrainedFV, Jin2021TowardsFR} that Gaussian kernel can summarize matching features effectively.
We construct a fine-grained translation matrix \( \mathbf{M} \) based on the cosine similarity, where \( \mathbf{M}_{ij} = cos(\mathbf{H}_{ei}, \mathbf{H}_{cj})\).
Subsequently, the Gaussian kernel is applied to transform the translation matrix into kernel features, attending to different levels of relevance. It can be denoted as:
\begin{align}
    \label{eq:score-function}
    & \mathbf{S}_{se} = \mathbf{\hat{H}}_{e}\mathbf{W}_{se} \\
    & \mathbf{K}_{i} = [\mathbf{K}_{i1}; \mathbf{K}_{i2}; \ldots; \mathbf{K}_{ik}] \\
    & \mathbf{K}_{it} = \log \sum_{j}{exp(- \frac{(\mathbf{M}_{ij} - \mu_t)^2}{2\sigma_t^2}) } \\
    & \mathbf{S}_{sc} = \mathbf{K} \mathbf{W}_{sc}
\end{align} 
where \( \mathbf{W}_{se} \in \mathbb{R}^{d \times 1}, \mathbf{W}_{sc} \in \mathbb{R}^{k \times 1}\) are trainable weights that project representations into the shared 1-dimension score space. \(k \) denotes the number of kernel with corresponding mean \( \mu_t\) and width \(\sigma_t \) that captures different specific similarity regions \cite{xiong2017end}. 

However, the redundancy is not only related to the information contained for claim verification in each node, but also induced by the contextual information, which is involved in the neighborhood on graphs. For example, if a claim can be verified by a snippet in an evidence, the rest of segments (including the snippet's context) will be redundant. Therefore, we utilize a 1-layer GGNN to compute context-aware redundancy scores, which takes into account both node-self and context information. Finally the two scores are fused by a simple linear combination. Mathematically, it can be formulated as:
\begin{align}
    \label{eq:ggnn-gsl}
    & \mathbf{s}_{se} = \textbf{GGNN}(\mathbf{\tilde{A}}, \mathbf{S}_{sc}) \\
    & \mathbf{s}_{sc} = \textbf{GGNN}(\mathbf{\tilde{A}}, \mathbf{S}_{se}) \\
    & \mathbf{s}_{r} = (1-\beta) \mathbf{s}_{se} + \beta \mathbf{s}_{sc} \\
    & idx = topk\_index(\mathbf{s}_r) \\
    & \mathbf{\tilde{A}}_{{idx}, :} = \mathbf{\tilde{A}}_{:, {idx}} = 0
    \label{eq:mask}
\end{align}
where \( \beta\) is an introduced coefficient that controls the fusion proportion of node-self and claim-related score. \(idx\) denotes the indices of node with top-\(k\) redundancy scores which are discarded by masking their degrees as 0 (c.f., Eq. (\ref{eq:mask})). Note that \(\textbf{GGNN}(\cdot)\) in Eq. (\ref{eq:ggnn-gsl}) does not share parameters with the semantics encoder due to their different targets. Besides, we only perform semantic structure refinement on evidences since claims are usually short (less than 10 words) so that the semantic structures are simple and unnecessary to be refined.

Notably, as the trainable parameters related to redundancy scoring need to be updated by back-propagation, we modify the Eq.(\ref{eq:ggnn-s}) in \(\textbf{GGNN}(\cdot)\) when it is just after semantic structure refinement by multiplying the normalized score to scale features:
\begin{align}
    \mathbf{a}_{i}&=\sum_{(w_{i}, w_{j}) \in \mathcal{C}} \mathbf{\tilde{A}}_{ij} \mathbf{W}_{a} \mathbf{H}_{j} (1-\sigma(\mathbf{s}_{rj}))
\end{align}
where \(\sigma\) is the Sigmoida function.

Finally, we stack the modified semantics encoder over one semantic structure refinement layer to form a unified module, namely \emph{semantic refinement and miner} (SRM in short), where the the redundant information is reduced and long-distance semantic dependency is captured based on refined structure. In general, we can first utilize a semantics encoder to perform neighborhood propagation, then stack \(T_R\) layers of SRM to refine the semantic structures \(T_R\) times, eventually obtaining the fine-grained representations.

\subsubsection{Attentive Graph Readout Layer}
\label{sec:interaction module}
So far, we have obtained refined structures \(\mathbf{\tilde{A}}^{(T_{R})}_{e}\) for each evidence and fine-grained node embeddings \(\mathbf{H}^{(T_{E})}_{c}\), \(\mathbf{H}^{(T_{R}+1)}_e\) for claims and evidences separately\footnote{We omit the index subscript of evidences for brevity, as they are all fed into the same networks.}, where \(T_{R}\) and \(T_{E}\) are the numbers of the SRM layer and semantics encoder layer of the claim, respectively (\(T_{R} = 1\) and \(T_{E} = 2\) in Figure \ref{fig:model}). Next, to perform the claim-evidence interaction, we first need to integrate all node embeddings (word embeddings) into general graph embeddings (claim and evidence embeddings). Following previous work \cite{vo2021hierarchical}, we propose to obtain claim-aware evidence representations via a word-level attentive layer. In detail, we compute the attention score of the \(j\)-th word \(\mathbf{H}_{ej}\) in the refined evidence graph with the claim representation \(\mathbf{h}^r_c\). Thereafter, the evidence representation \(\mathbf{h}^{r}_{e}\) is obtained via weighted summation:

\begin{align}
    \mathbf{h}^r_c &= \frac{1}{l_c} \sum_{i=1}^{l_c} \mathbf{H}_{ci} \\
    \label{eq:attn-s}
    \mathbf{p}_j &= \tanh \left(\left[\mathbf{H}_{ej} ; \mathbf{h}^r_c\right] \mathbf{W}_{c}\right) \\ 
    \alpha_j &= \frac{\exp \left(\mathbf{p}_j \mathbf{W}_{p}\right)}{\sum_{i=1}^{l_e} \exp \left(\mathbf{p}_i \mathbf{W}_{p}\right)} \\
    \mathbf{h}^{r}_{e} &= \sum_{j=1}^{l_{e}} \alpha_{j} \mathbf{H}_{ej}
    \label{eq:attn-e}
\end{align}
where \([\cdot; \cdot]\) denotes the concatenation of two vectors and \(\mathbf{W}_c \in \mathbb{R}^{2d \times d}\) and \(\mathbf{W}_p \in \mathbb{R}^{d \times 1}\) are the trainable parameters. \(l_c\) and \(l_e\) are the length of claim and evidence, respectively. We denote Eq. (\ref{eq:attn-s}) - (\ref{eq:attn-e}) as \(\textbf{ATTN}(\mathbf{H_e}, \mathbf{h}^r_c)\) and the attention modules can be easily extended to multi-head ones by concatenating outputs of each head. 
% It is worth noting that based on the fine-grained representations our graph-based model outputs, the above attention mechanism can be replaced by any interaction method in previous work.

As MAC \cite{vo2021hierarchical} empirically demonstrates that claim speaker and evidence publisher information is important for verification, we extend claim and evidence representations by concatenating them with corresponding information vectors:
\begin{align}
    \mathbf{h}_c &= [\mathbf{h}^{r}_c; \mathbf{s}]\\
    \mathbf{h}^{g}_e &= [\mathbf{h}^{r}_e; \mathbf{p}]
\end{align}

After obtaining the claim and evidence representations, we further employ another document-level attentive layer, which is of the same structure as the above, to capture the document-level interaction between a claim and several evidences:

\begin{align}
    \mathbf{H}^{g}_e &= [\mathbf{h}^{g}_{e1}; \mathbf{h}^{g}_{e2}; \ldots; \mathbf{h}^{g}_{en}] \\
    \mathbf{h}_e &= \textbf{ATTN}(\mathbf{H}^{g}_e, \mathbf{h}_c)
    \label{2nd_attn}
\end{align}
where \(\mathbf{H}^{g}_e\) denotes the concatenation of embeddings of \(n\) evidences. Eventually, we integrate claim and evidence embeddings into one unified representation via concatenation, followed by a multi-layer perceptron to output the veracity prediction \(\hat{y}\). 
\begin{align}
    \mathbf{h} &= [\mathbf{h}_c; \mathbf{h}_e] \\
    \hat{y} &= \text{Softmax}(\mathbf{W}_f \mathbf{h} + \mathbf{b}_f)
\end{align}

As it is fundamentally a classification task, we utilize the standard cross entropy loss \(\mathcal{L}_{ce}\) as the task loss, which can be written as:
\begin{align}
    \mathcal{L}_{ce}  =- (y\log \hat{y}+(1-y) \log (1-\hat{y}))
\end{align}
where \(y \in \{0, 1\}\) denotes the label of each unverified news.

\subsubsection{Adversarial Contrastive Learning Module}
% 1.problem 2. advantage of using contrastive learning 3. method

% As the abundant relations between samples of different classes imply the proximity of samples in the representation space, we propose an auxiliary supervised contrastive learning task to calibrate the representation, in addition to the veracity classification task. The supervised contrastive learning aims to pull samples of the same class close while pulling samples of different classes apart, so as to obtain separable representations that can capture key clues of different claim-evidence interactions. 

To make full use of data, we propose an auxiliary supervised contrastive learning task to help calibrate representations. The supervised contrastive learning aims to pull samples of the same class close and samples of different classes apart, exploiting the common properties within the class. Corresponding contrastive auxiliary loss \(\mathcal{L}_{sup}\) takes the following form:
\begin{gather}
    % \begin{split}
      \mathcal{L}_{cl}={\frac{-1}{|\mathcal{P}(\mathbf{h})|}\sum_{\mathbf{h}_p \in \mathcal{P}(\mathbf{h})}\log\frac{\exp (cos(\mathbf{h}, \mathbf{h}_p) /\tau)}{\sum_{\mathbf{h}_n\in \mathcal{N}(\mathbf{h})}\exp(cos(\mathbf{h}, \mathbf{h}_{n}) /\tau) }}
    %  \end{split}
\end{gather}
where \( \mathcal{P}(\mathbf{h})\) is the set of positive samples and \(\mathcal{N}(\mathbf{h})\) is the set of negative samples to \(\mathbf{h}\) in the same batch. \(\mathbf{h}_p\) is the positive sample with the same label and \(\mathbf{h}_n\) is the negative sample with a different label. \(\tau \in \mathbb{R} \) is a temperature parameter and \(cos(\cdot)\) denotes the cosine similarity function.

Moreover, to further reduce the sensitivity of models to local evidence, we employ adversarial gradient perturbation  \cite{miyato2016adversarial, li2022hiclre} to construct augmented samples for more efficient contrastive learning. It can be viewed as an intentional injected noise. Compared to discrete augmentations on textual content like word deletion, insertion, and substitution which may hurt the semantics of complex sentences, the gradient-based method can maintain semantics to the greatest extent and simulate a worst case. These augmented instances are integrated to enrich the representations in the representation space. 

Specific to the augmentation process, we first select the piece of evidence \(\mathbf{h}^{g}_{ek}\) with the highest attention score in \(\mathbf{H}_{e}^{g}\), which contributes most to the synthetic evidence representation \(\mathbf{h}_e\).
Then we utilize the gradient adversarial perturbation to \(\mathbf{h}^{g}_{ek}\) in the representation level to obtain the perturbed evidence \(\mathbf{h}^{g'}_{ek}\). Finally, the perturbed evidence along with other evidence representations is input into the same document-level attentive network  \(\textbf{ATTN}()\) to obtain the final augmented view \(\mathbf{h}^{'}\). So far, the positive set \(\mathcal{P}(\mathbf{h}) \) contains the original samples and adversarial augmented samples of the same class, and the negative set \(\mathcal{N}(\mathbf{h})\) contains those of different classes. The perturbation process above can be mathematically expressed as:
\begin{align}
    \mathbf{g}_{e k} &= \nabla_{\mathbf{h}^{g}_{e k}}\mathcal{L}_{ce} \\
    \mathbf{h}^{g'}_{e k} &= \mathbf{h}^{g}_{e k} + \epsilon \frac{\mathbf{g}_{e k}}{||\mathbf{g}_{ek}||} \\
    \mathbf{H}^{g'}_e &= [\mathbf{h}^{g}_{e1}; \ldots; \mathbf{h}^{g'}_{e k}; \ldots] \\
    \mathbf{h}^{'}_e &= \textbf{ATTN}(\mathbf{H}^{g'}_e, \mathbf{h}_c) \\
    \mathbf{h}^{'} &= [\mathbf{h}_c; \mathbf{h}^{'}_e]
\end{align}
where \(\mathbf{g}_{ek} \) is the first-order derivation of target loss at variable \(\mathbf{h}^{g}_{ek}\). \(\epsilon\) is the norm parameter to control the normalized gradient as a valid perturbation. Then \(\mathbf{h}^{'}\) is the unified representation of an augmented view that shares the same label with the original sample. These instances are used to simulate the instances with noise, thus improving the difficulty of contrastive learning.

\subsubsection{Training Objective}
Finally, we combine the cross entropy loss and supervised contrastive loss as a joint optimization target loss:
\begin{gather}
    \mathcal{L}  = \mathcal{L}_{ce} + \lambda \mathcal{L}_{cl}
\end{gather}
where \(\lambda\) is a hyper-parameter to control the extent of contrastive learning.

\begin{table}[]
    \caption{The statistics of two datasets. The symbol ``\#'' denotes ``the number of''. ``True'' and ``False'' stand for true claims and false claims, respectively. ``Evi.', `Spe.'', and ``Pub.'' denote evidences, speakers and publishers.}
  \resizebox{0.48\textwidth}{!}{
    \footnotesize
    \begin{tabular}{cccccc}
        \toprule
        Dataset & \# True & \# False & \# Evi. & \# Spe. & \# Pub. \\
        \midrule
        Snopes & 1164  & 3177  & 29242 & N/A   & 12236 \\
        PolitiFact & 1867  & 1701  & 29556 & 664   & 4542 \\
        \bottomrule
    \end{tabular}
  }
  \label{tab:dataset}
\end{table}

\begin{table*}[htbp]
     \caption{The model comparison on two datasets Snopes and PolitiFact. ``F1-Ma'' and ``Fi-Mi'' denote the metrics F1-Macro and F1-Micro, respectively. ``-T'' represents ``True News as Positive'' and ``-F'' denotes ``Fake news as Positive'' in computing the precision and recall values. The best performance is highlighted in boldface. \(\ddag\) indicates that the performance improvement is significant with p-value \(\leq\) 0.05.}
    \centering
    \resizebox{\textwidth}{!}{
    \begin{tabular}{ccccccccc|cccccccc}
    \hline
    \multirow{2}{*}{Method} & \multicolumn{8}{c|}{Snopes}                                     & \multicolumn{8}{c}{PolitiFact} \\
    \cline{2-17}
    & F1-Ma & F1-Mi & F1-T    & P-T & R-T & F1-F    & P-F  & R-F & F1-Ma & F1-Mi & F1-T    & P-T & R-T & F1-F    & P-F     & R-F  \\
    \hline
    % & 0.561    & 0.698      & 0.316     & 0.403    & 0.261   & 0.806   & 0.760   & 0.858 & 0.560     & 0.564    & 0.564      & 0.592   & 0.550      & 0.557   & 0.544   & 0.581 
    LSTM    & 62.10    & 71.87      & 42.95     & 48.42    & 39.69   & 81.25   & 79.14   & 83.67  & 60.56   & 60.87    & 61.82      & 63.19   & 61.27      & 59.31   & 59.05   & 60.43 \\
    TextCNN & 63.08    & 72.01      & 45.00     & 48.16    & 43.04   & 81.16   & 79.88   & 82.62 & 60.38   & 60.74    & 61.52      & 63.01   & 61.03      & 59.24   & 59.05   & 60.42 \\
    BERT & 62.05	& 71.62	& 43.07	& 47.73	& 40.65	& 81.04	& 79.31	& 82.97  & 59.71	& 59.81	& 60.81	& 61.95	& 59.90	& 58.60	& 57.73	& 59.70  \\
    % BERT & 62.1	& 71.6	& 43.1	& 47.7	& 40.7	& 81.0	& 79.3	& 83.0  & 59.7	& 59.8	& 60.8	& 61.9	& 59.9	& 58.6	& 57.7	& 59.7  \\
    \hline
    DeClarE & 72.54 & 78.61      & 59.43    & 61.03    & 57.93    & 85.67   & 85.25  & 86.39  & 65.31   & 65.25     & 67.49    & 66.71   & 68.32      & 63.11   & 63.70   & 62.46 \\
    HAN     & 75.21      & 80.23      & 63.58    & 62.50    & 64.69    & 86.83   & 87.64  & 86.11 & 66.12     & 66.01     & 67.92   & 67.58   & 68.20      & 64.33  & 64.97   & 63.73 \\
    EHIAN   & 78.43      & 82.83      & 68.41    & 61.69    & 76.79    & 88.47   & 88.18  & 89.04  & 67.22     & 67.95     & 68.92    & 68.64   & 69.34      & 65.52   & 67.49   & 63.60 \\
    MAC     & 78.66      & 83.32      & 68.74    & 70.00    & 68.60    & 88.58   & 88.62  & 88.71 & 68.03     & 68.25     & 71.78    & 67.54   & 73.49      & 64.28   & 67.61   & 61.68 \\
    CICD    & 78.92      & 83.73      & 69.07    & 63.20    & \textbf{77.48}    & 89.30   & \textbf{88.99}  & 89.54   & 68.18     & 68.48     & 70.24    & 68.92   & 71.44      & 65.72   & 69.12   & 62.93 \\
    \hline
    % \themodel     & \(\textbf{0.8030}^\ddag\)      & \(\textbf{0.8468}^\ddag\)      & \(\textbf{0.7114}^\ddag\)    & \(\textbf{0.7178}^\ddag\)    & 0.7092    & \(\textbf{0.8946}^\ddag\)   & \textbf{0.8941}  & \(\textbf{0.8958}^\ddag\)  & \(\textbf{0.6951}^\ddag\)    & \(\textbf{0.6978}^\ddag\)    & \(\textbf{0.7208}^\ddag\) & 0.6979 & \(\textbf{0.7474}^\ddag\)      & \(\textbf{0.6693}^\ddag\)  & \(\textbf{0.7008}^\ddag\)  & 0.6433 \\
    \themodel     & \(\textbf{80.61}^\ddag\)      & \(\textbf{85.12}^\ddag\)      & \(\textbf{71.26}^\ddag\)    & \(\textbf{74.18}^\ddag\)    & 68.79    & \(\textbf{89.96}^\ddag\)   & \(88.90\)   & \(\textbf{91.04}^\ddag\)  
    & \(\textbf{69.53}^\ddag\)    & \(\textbf{69.81}^\ddag\)    & \(\textbf{72.21}^\ddag\) & \(\textbf{69.73}^\ddag\) & \(\textbf{75.10}^\ddag\)      & \(\textbf{66.84}^\ddag\)  & \(\textbf{70.26}^\ddag\)  & \(\textbf{64.01}^\ddag\) \\
    \hline
    \end{tabular}
    }
  \label{tab:comp}
\end{table*}%

\section{experiments}
\label{experiments}
In this section, we conduct comprehensive experiments to answer the following research questions:
\begin{itemize}
    \item RQ1: How does \themodel perform compared to previous fake news detection baselines?
    \item RQ2: How effective are the structure modeling component and contrastive learning component to \themodel?
    % \item RQ3: How robust \themodel is when tested in noisy scenes?
    \item RQ3: How does \themodel perform under different hyperparameter settings?
\end{itemize}

\subsection{Experimental Setup}
\subsubsection{Datasets}
We utilize two widely used datasets to verify our proposed model. The detailed statistics is summarized in Table \ref{tab:dataset}.
\begin{itemize}
    \item Snopes \cite{Popat2017WhereTT}. Claims and their corresponding labels (\(true\) or \(false\)) are collected from the fack-checking website\footnote{https://www.snopes.com/}. Taking each claim as a query, the evidences and their publishers are retrieved via the search engine.
    \item PolitiFact \cite{vlachos-riedel-2014-fact}. Claim-label pairs are collected from another fact-checking website\footnote{https://www.politifact.com/} about US politics and evidences are obtained in a similar way to that in Snopes. Aside from publisher information, claim promulgators are added into the dataset. Following previous work \cite{Rashkin2017TruthOV, popat2018declare, vo2021hierarchical}, we merge \(true\), \(mostly\) \(true\), \(half\) \(true\) into the unified class \(true\) and \(false\), \(mostly\) \(false\), \(pants\) \(on\) \(fire\) into \(false\). 
\end{itemize}

\subsubsection{Baselines}
To demonstrate the effectiveness of our proposed model \themodel, we compare it with several existing methods, including both pattern- and evidence-based models, and the specific description is listed as follows:

\textbf{Pattern-based methods.}
\begin{itemize}
    \item \textbf{LSTM} \cite{lstm} utilizes LSTM to encode the semantics with the news as input and obtains the final representation of claim via average pooling. 
    \item \textbf{TextCNN} \cite{textcnn} applies a 1D-convolutional network to embed the semantics of claim.
    \item \textbf{BERT} \cite{bert} employs BERT to learn the representation of claim. A linear layer is stacked over the special token [CLS] to output the final prediction. 
\end{itemize}

\textbf{Evidence-based methods.}
\begin{itemize}
    \item \textbf{DeClarE} \cite{popat2018declare} employs BiLSTMs to embed the semantics of evidences and obtains the claim's representation via average pooling, followed by an attention mechanism performing among claim and each word in evidences to generate the final claim-aware representation.
    \item \textbf{HAN} \cite{ma2019sentence} uses GRUs to embed semantics and designs two modules named topic coherence and semantic entailment to model the claim-evidence interaction, which are based on sentence-level attention mechanism. 
    \item \textbf{EHIAN} \cite{wu2020evidence} utilizes self-attention mechanism to learn semantics and concentrates on the important part of evidences for interaction.
    \item \textbf{MAC} \cite{vo2021hierarchical} introduces a hierarchical attentive framework to model both word- and evidence-level interaction.
    \item \textbf{CICD} \cite{wu2021unified} introduces individual and collective cognition view-based interaction to explore both local and global opinions towards a claim.
\end{itemize}

\subsubsection{Implementation Details}
% We introduce the specific settings in our experiments including hyperparameters, training settings, and the experimental environment.

Following previous work \cite{popat2018declare, vo2021hierarchical}, we utilize the same data split\footnote{https://github.com/nguyenvo09/EACL2021/tree/main/format-ted\_data/declare} to train and test our model. We also report 5-fold cross validation results, where 4 folds are used for training and the rest one fold is for testing. We utilize Adam optimizer with a learning rate \(lr=0.0001\) and weight decay \(decay=0.001\). The model early stops when F1-macro does not increase in 10 epochs and the maximum number of epoch is 100. We set the maximum length of claims and evidences in both datasets as 30 and 100, respectively. The number of evidences \(n = 30\) and the batch size is 32. The fusion rate \(\beta\) is \(0.5\). The kernel size is set to 11 for Snopes and 21 for PolitiFact. Specifically, one kernel with \(\mu=1\) and \(\sigma=10^{-3}\) can capture exact matches \cite{Jin2021TowardsFR}, while \(\mu\) of the other kernels is spaced evenly between \([-1, 1]\) and \(\sigma\) is set to 0.1. We set the redundancy discarding rate \(r = 0.3\) for Snopes and \(r = 0.2\) for PolitiFact, i.e., \(k = rl_e\) will be filtered out in a semantic refinement layer, where \(l_e\) is the length of evidence. The number of semantics encoder layer \(T_E\) and evidence semantics miner layer \(T_R \) is both 1. The number of word-level and document-level attentive readout head as 5 and 2 for Snopes (3 and 1 for PolitiFact), the dimension of publisher and speaker embedding is both 128, following the work \cite{vo2021hierarchical}. We use the Glove pretrained embedding with the dimension \(d=300\) for all baselines for a fair comparison. We conduct all experiments using PyTorch 1.5.1 on a Linux server equipped with GeForce RTX 3090 GPUs and AMD EPYC 7742 CPUs.

\subsection{Model Comparison (RQ1)}
We compare our model \themodel with eight baselines\footnote{As some evidence-based methods do not release codes, we reproduce results carefully following settings reported in their original publications.}, including three pattern-based methods and five evidence-based methods. The overall results are shown in Table \ref{tab:comp}, from which we have the following observations:

Firstly, our model \themodel outperforms all existing methods on most of metrics on both two datasets by a significant margin, demonstrating the effectiveness of \themodel. It is worth noting that \themodel stands out from the recent three sequential-based baselines (EHIAN, MAC, and CICD) whose performances are close, indicating the positive impact of introducing graph-based models and contrastive learning to evidence-based fake news detection. In detail, compared to the strongest baselines CICD on two datasets, \themodel advances the performance by about 1.5 percent on F1-Macro and F1-Micro, which can better reflect the overall detection capability of models. With regard to the more fine-grained evaluation, i.e., `True news as Positive' and `Fake news as Positive', \themodel also achieve the best results on the F1 score on two datasets, where the F1 score is more representative than Precision and Recall since it takes into account both of them. 

Secondly, compared to the pattern-based methods (i.e., the first three methods in Table \ref{tab:comp}), evidence-based approaches have a substantial performance improvement. This is probably due to the better generalization of evidence-based methods, where the external information is utilized to probe the claim veracity, avoiding the over-reliance on text patterns.
In addition, the performance of BERT is similar to that of other pattern-based approaches. We suspect the reason is probably that claims are short and contain lots of noises (e.g., spelling errors and domain-specific abbreviations), which are rarely appeared in the pretraining corpus, thus it is hard for BERT to transfer the contextual information learned from the pretrained stage. 
% Even so, most of evidence-based models surpass BERT, indicating the advantage and necessity of incorporating evidences in fake news detection.

Thirdly, among five evidence-based baselines, the performance of DeClarE and HAN is inferior to other three models, which is mainly because they lack exploring the different grain-sized semantics. Specifically, DeClarE only considers word-level semantic interaction and HAN solely relies on document-level representations to model claim-evidence interaction. However, the rest of evidence-based methods all consider multi-level semantics, thus achieving better performance.

\subsection{Ablation Study (RQ2)}
% \textbf{RQ2: How does the redundant information involved in evidences affect the model performance?}

\begin{table}[t]
  \caption{The performance comparison between \themodel and model variants.}
  \centering
    \begin{tabular}
    {c c c  | c c }
    	\toprule
        \multirow{2}{*}{Method} 
        & \multicolumn{2}{c|}{Snopes}  
        & \multicolumn{2}{c}{PolitiFact} 
        \\
        \cline{2-5}
        & F1-Ma & F1-Mi & F1-Ma & F1-Mi \\
        \hline
        \themodel-SE-CL & 77.51 & 82.31 & 67.47 & 67.77 \\
        \themodel-GSE-CL  & 78.66 & 83.32 & 68.03 & 68.25 \\
        \themodel-SSR-CL  & 79.49 & 84.10 & 68.45 & 68.78 \\
        % & 79.16 & 83.79 & 68.19 & 68.66
        \themodel-CL  & 80.12 & 84.52 & 69.25 & 69.60 \\
        \themodel-AD & 80.32 & 84.81 & 69.40 & 69.69 \\
        \hline
        \themodel & 80.61 & 85.12 & 69.53 & 69.81  \\
    \bottomrule
    \end{tabular}%
  \label{tab:ablation}%
\end{table}%

To verify the effect of components related to fine-grained semantics mining or contrastive learning of \themodel, we conduct the ablation study for several variants by removing the specific component: \textbf{-SE} removes any semantics encoder and feeds the pretrained word embeddings e.g. Glove, into the attentive readout layer directly; \textbf{-GSE} removes the graph semantic encoder and utilizes the BiLSTM as the semantics encoder like the baseline \cite{vo2021hierarchical}; \textbf{-SSR} removes the structure learning layer which is proposed to reduce the useless redundancy in evidences; \textbf{-CL} only uses the task specific classification loss \(\mathcal{L}_{ce}\) for training; \textbf{-AD} performs a simplified supervised contrastive learning without constructing the augmented instances by adversarial gradient perturbation.

The experimental results are shown in Table \ref{tab:ablation}, from which we can observe that each variant suffers form an obvious decline on both datasets regarding the F1-Micro and F1-Macro. Specific analyses are as follows:

\begin{itemize}
    \item In terms of different semantics encoders, \themodel-SE-CL has the poorest performance since the contextual information is not captured. Moreover, the performance of \themodel-SSR-CL is superior to that of \themodel-GSE-CL, indicating that the long-distance structural dependency involved in semantic structure, which is less explored in sequential models, is significant for veracity checking. Note that we choose \themodel-SSR-CL instead of \themodel-CL to be compared with \themodel-GSE-CL fairly, since the only difference between \themodel-SSR-CL and \themodel-GSE-CL is the semantics encoder.
    \item  The performance degradation of \themodel-SSR-CL demonstrates the necessity of performing structure refinement on semantic graphs and confirms the effectiveness of our structure learning method. Furthermore, it indicates that reducing the effect of unimportant information via attention mechanisms will lead to suboptimal results, since it still maintains the noisy semantic structure unchanged \cite{chen2020gsl} (i.e., specifically, all words will participate in the claim-evidence interaction). Therefore, the effect of structure refinement is not overlapped with the attention mechanism, but further goes beyond.  
    \item Additionally, \themodel gains significant improvements on \themodel-CL and \themodel-AD. This indicates that contrastive learning and adversarial gradient perturbation are both beneficial. Contrastive learning can mine the underlying relations to help capture the core difference between classes, while the adversarial augmented instances can further boost this process. It is worth noting that gradient adversarial augmentation has a more significant effect on the Snopes dataset compared to PolitiFact. We attribute this improvement to the supplement of obviously unbalanced samples in Snopes, which promotes efficient contrastive learning.
\end{itemize}

\subsection{Sensitivity Analysis (RQ3)}
In this section, we conduct experiments to analyse the performance fluctuation of \themodel with respect to different values of key hyperparameters. 

\subsubsection{The number of semantics encoder layer for claims \(T_E\)} 
\begin{figure}[t]
   \begin{center}
   \includegraphics[width=0.45\textwidth]{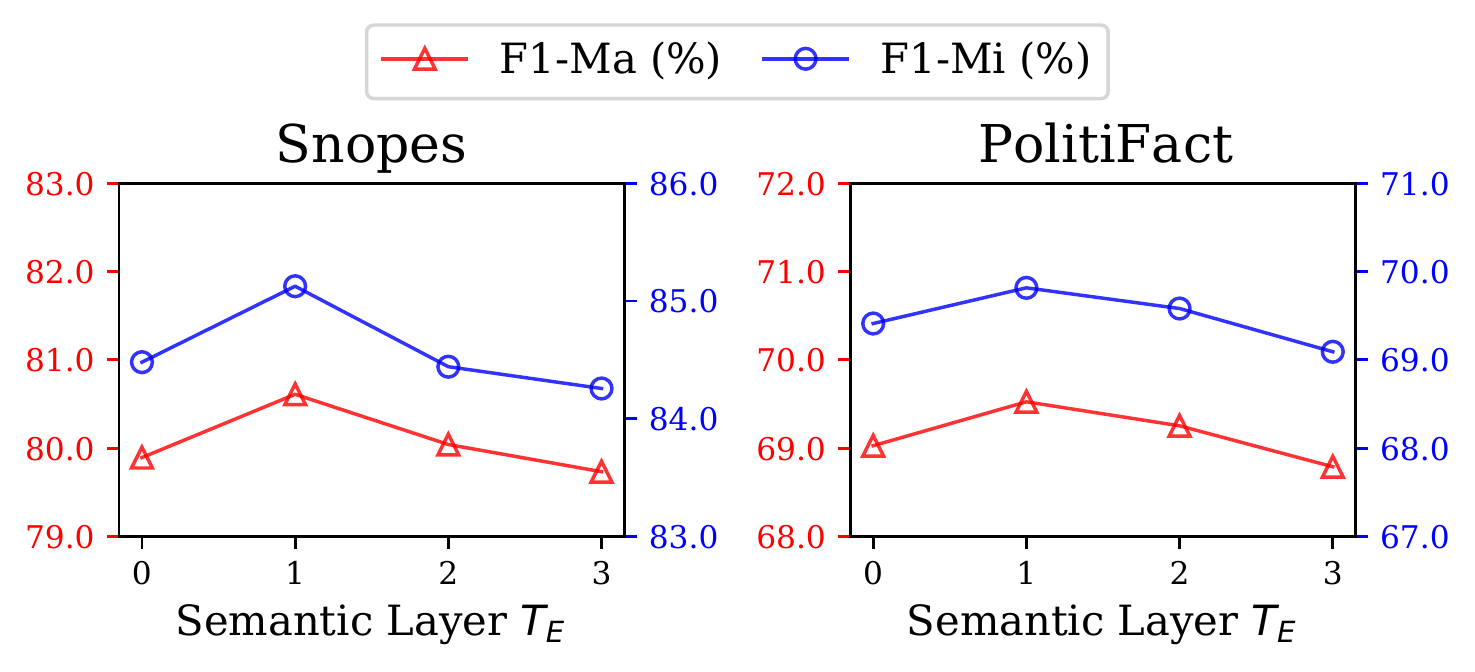}
   \end{center}
   \caption{The influence of different semantics encoder layers \(T_E\) for claims on model performance.}
   \label{fig:ggnn_num}
\end{figure}

This hyperparameter decides the propagation field on graphs, since stacking \(T_E\)-layer encoder (GGNN) makes each node aggregate information within \(T_E\)-hop neighborhood. We report the model performance when \(T_E = 0, 1, 2, 3\) (See Figure \ref{fig:ggnn_num}) and summarize the observations as follows:

There is a significant improvement when \(T_E\) is changed from 0 to 1. Specifically, the model with \(T_E = 1\) outperforms its counterparts. We suspect that the gains are due to the short length of claims (the average lengths of claim are about 6 and 8 in Snopes and PolitiFact, respectively), where the semantic structure can be well-explored merely via 1-hop propagation.

An obvious decline are observed between \(T_E=1\) and \(T_E=3\), which is probably caused by the inappropriate propagation field. When the layer number is increased, each node on graphs aggregates information from the multi-hop neighborhood, which may cover all nodes since the claims are short, thus failing to model the local semantic structure and leading to poor performance.

% There is no drastic rise and fall when \(T_E\) is changed from 0 to 3. Specifically, the model with \(T_E = 1\) outperforms its counterparts. We suspect that the close results are due to the short length of claims (the average lengths of claim are about 6 and 8 in Snopes and PolitiFact, respectively), where the semantic structure can be well-explored merely via 1-hop propagation.

% Only one obvious decline is observed between \(T_E = 2\) and \(T_E = 3\), which is probably caused by the inappropriate propagation field. When the layer number is 3, each node on graphs aggregate information from 3-hop neighborhood, which may cover all nodes since the claims are short, thus failing to model the local semantic structure and leading to the poor performance.

\begin{figure}[t]
   \begin{center}
   \includegraphics[width=0.45\textwidth]{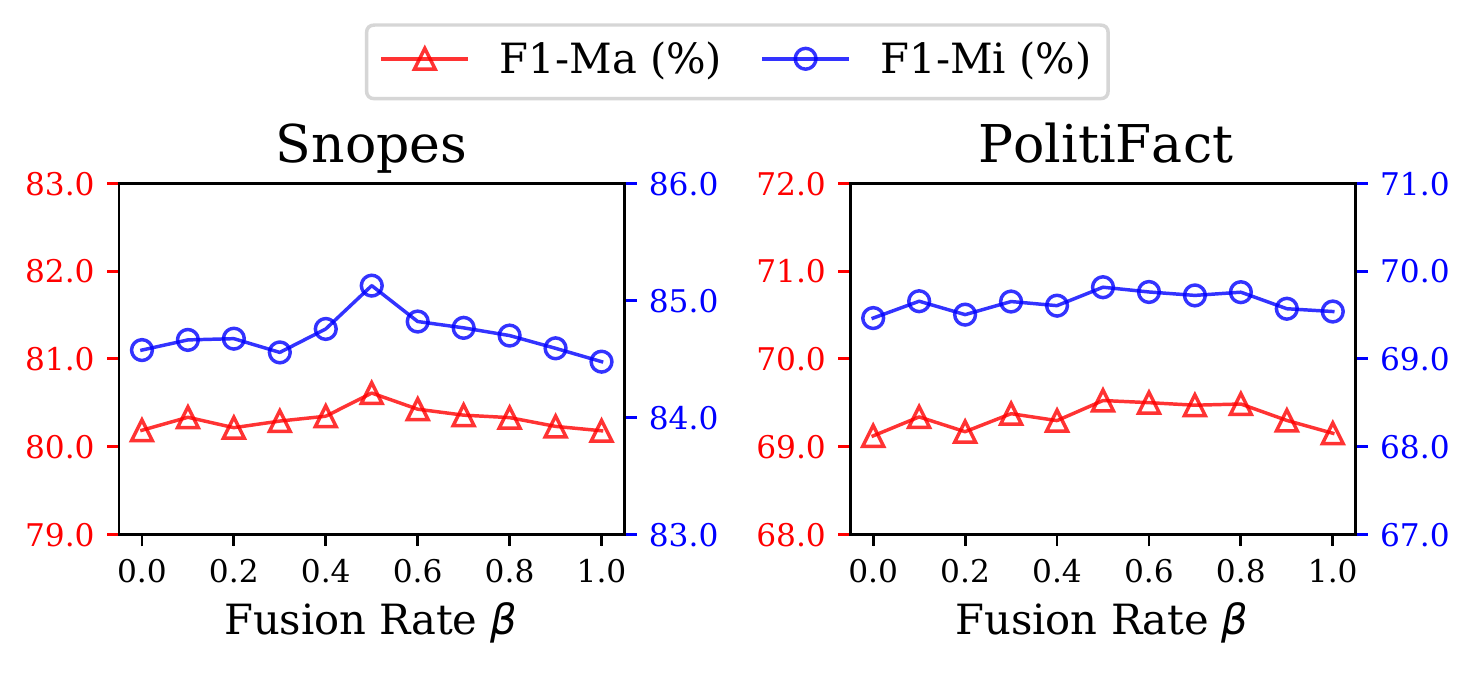}
   \end{center}
   \caption{The influence of different fusion rate \(\beta\) on model performance.}
   \label{fig:fusion_rate}
\end{figure}

\subsubsection{The fusion coefficient \( \beta \)}
The fusion coefficient \(\beta\) controls the fusion rate of node-self and claim related score in the graph structure learning. \(\beta=0 \) denotes \themodel only considers the information within evidence to determine redundancy, and \(\beta=1\) denotes \themodel only considers the fine-grained claim-related information to determine redundancy. From the results in Figure \ref{fig:fusion_rate}, we can observe that:

% performance gain when combine two scores, best score
% efficacy and insensitivity
When \(\beta\) is set to \(0\) or \(1\), the performance is relatively poor. Because it only determines redundancy from a single perspective, failing to exploit the rich information contained in each node and its relation to the claim. And when \(\beta\) changing from 0 to 1, the performance increases first and then decreases, indicating that a moderate fusion score is useful. It is worth noting that the best performance is achieved when \(\beta=0.5\). It denotes that the claim-related redundancy score is as important as the node-self score and they both should be taken into consideration.

With varied fusion coefficient \(\beta\),  \themodel always obtains a competitive performance on both datasets. It indicates that our method is relatively insensitive to the change of \(\beta\) and proves the validity of the fused redundancy score.

\subsubsection{The discarding rate \(r\)} 
\begin{figure}[t]
   \begin{center}
   \includegraphics[width=0.45\textwidth]{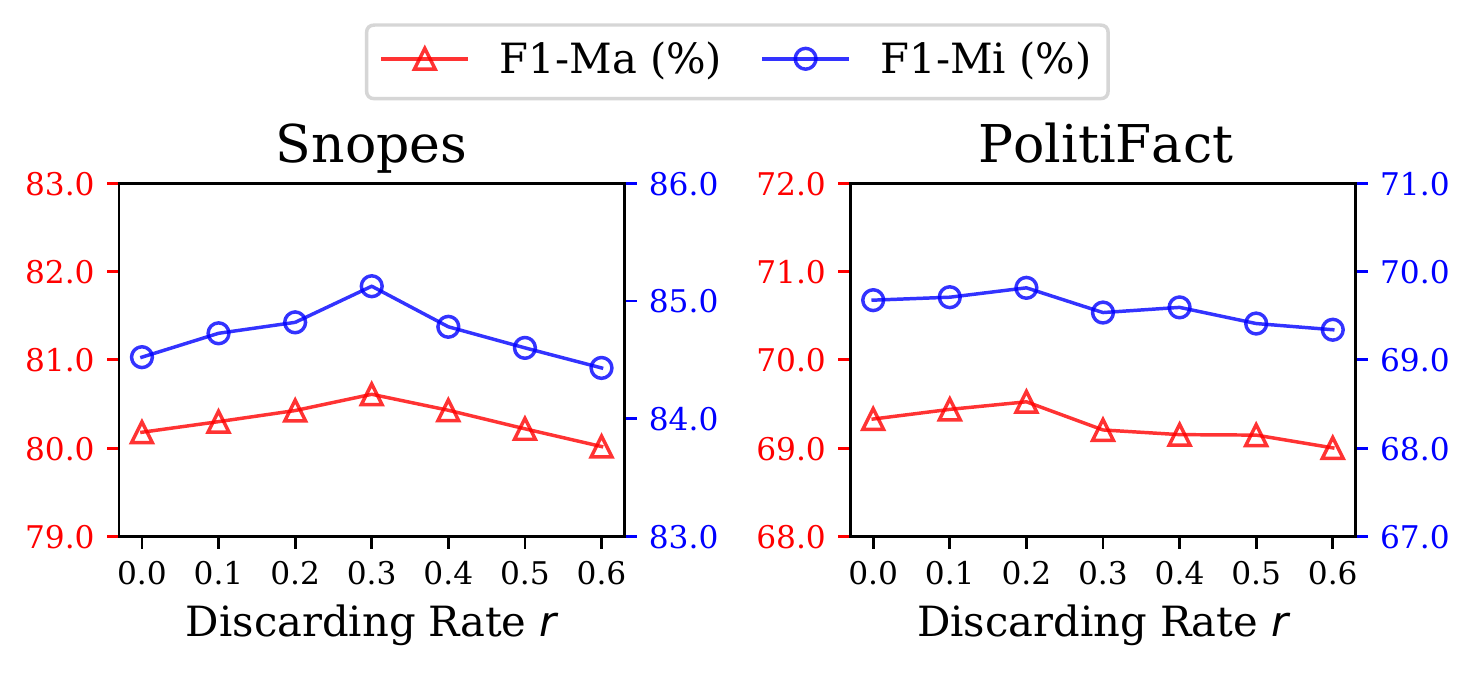}
   \end{center}
   \caption{The influence of different discarding rates \(r\) on model performance.}
   \label{fig:gsl_rate}
\end{figure}

This discarding rate decides the proportion of redundant information in evidences we filter out. We test the model with \(r\) ranging from 0 to 0.6 (See Figure \ref{fig:gsl_rate}) and have the following observations:

When \(r = 0\), the model can be viewed as integrating the redundancy signals when encoding semantics, without dropping nodes. It is worth noting that this degraded version of \themodel differs from the one without semantic structure refinement and can still maintain competitive performance. It is mainly because the former can adaptively reduce the weight of redundant nodes while the latter can't distinguish the difference and treat each node equally. 

The performance grows with \(r\) increasing and peaks at the best when \(r = 0.3\) in Snopes and \(r=0.2\) in PolitiFact, which indicates that reducing redundant information plays a positive role in improving the model performance. When \(r\) continues to increase, an obvious performance decline can be seen. The probable reason is that some useful information for veracity prediction is mistakenly discarded, so that the model fails to capture the rich semantics in evidences, as the \(r\) is too large.

\begin{figure}[t]
   \begin{center}
   \includegraphics[width=0.45\textwidth]{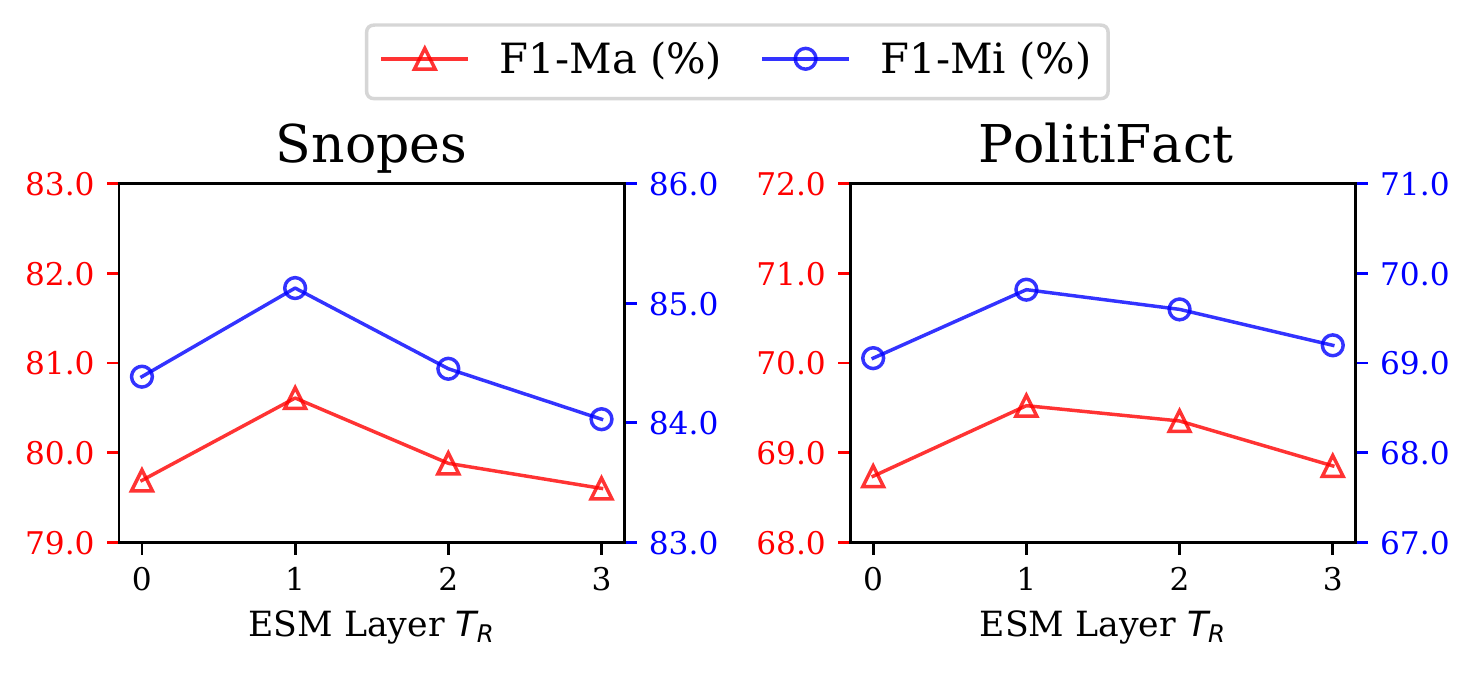}
   \end{center}
   \caption{The influence of different semantic refinement and miner layers \(T_R\) on model performance.}
   \label{fig:gsl_num}
\end{figure}

\subsubsection{The number of SRM layer \(T_R\)} 
It is a key hyperparameter that controls the information propagation field on graphs and the extent of structure refinement. We observe some phenomena when \(T_R\) increases from 0 to 3 (See Figure \ref{fig:gsl_num}):

The performance is first improved from \(T_R = 0\) to \(T_R = 1\). Note that when \(T_R = 0\), the model downgrades into the one with only a semantics encoder layer. The inferior performance is mainly due to two aspects: 1) It is unable to capture the high-order semantics of long evidences since only features from 1-hop neighborhood are aggregated. 2) Moreover, no redundancy reduction may affect other claim-relevant useful information, since this redundant information are fused via neighborhood propagation. Therefore, these drawbacks, in turn, demonstrate the significance of high-order semantics and structure refinement.

A significant fall of performance can be seen when \(T_R\) ranges from 1 to 3. This is probably because the networks suffer from the over-smoothing problem, which is common in GNNs \cite{Li2018DeeperII}. Besides, the information is overly discarded so that the evidence semantics is not well modeled.

\subsubsection{The contrastive coefficient \(\lambda \)}

\begin{figure}[t]
   \begin{center}
   \includegraphics[width=0.45\textwidth]{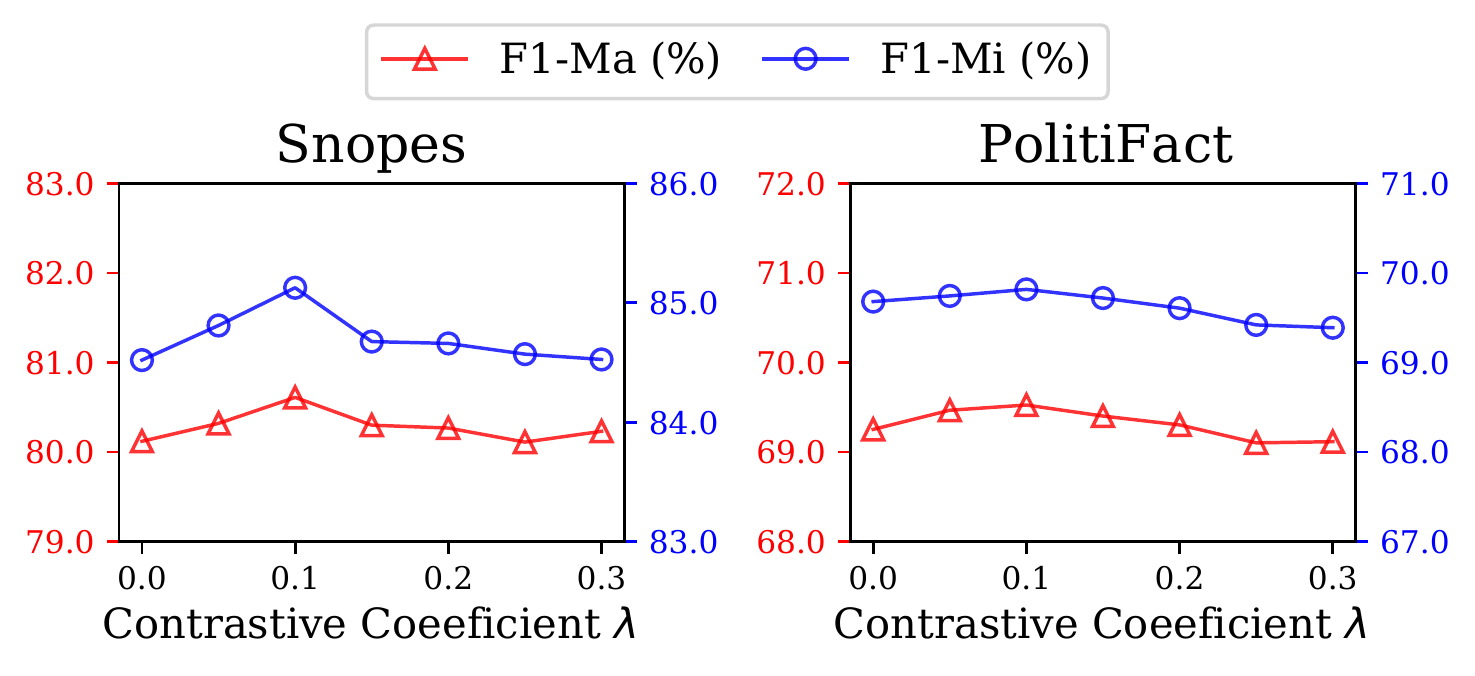}
   \end{center}
   \caption{The influence of different contrastive coefficient \(\lambda\) on model performance.}
   \label{fig:con_lambda}
\end{figure}

We also conduct experiments to study the impact of contrastive coefficient \(\lambda\), with different values ranging from 0.0 to 0.3 (See Figure \ref{fig:con_lambda}). This hyperparameter decides the extent of the auxiliary contrastive learning task besides classification task. We have the following observations:

% 1.first improved 2.decline
There is a significant improvement when \(\lambda\) ranging from \(0.0\) to \(0.10\), and the model achieves the best performance when \(\lambda=0.1\). In specific, \(\lambda=0.0\) denotes the simplified version \themodel-CL which only focuses on the classification task. It indicates that the auxiliary contrastive learning task can improve the performance of our method.

However, the performance begins to decline obviously when \(\lambda\) continues to increase. This is due to a moderate \(\lambda\) can help the model to capture separable and reliable representations which are beneficial to detection, while a larger value may distract it from the main task. The problem of sensitivity to the proportion of the auxiliary contrastive learning task has recently been noticed by existing works in \cite{zhang2021latent, yu2021self}.

\section{conclusion}
In this paper, we have proposed a unified graph-based fake news detection model with adversarial contrastive learning named \themodel to explore the complex semantic structure and enhance the representation learning. Based on constructed claim and evidence graphs, the long-distance semantic dependency is captured via the information propagation. Moreover, a simple and effective structure learning module is introduced to reduce the redundant information, obtaining fine-grained semantics that are more beneficial for the downstream claim-evidence interaction. Finally, we integrate an adversarial contrastive learning task to capture separable representations to help fake news detection. We have conducted empirical experiments to demonstrate the superiority of our proposed method.
% We have also validated the performance of \themodel with different interaction methods, where results demonstrate its ability of acting as a plug-in-play module to boost the performance of other fake news detection models.
% \input{section/appendix}
\ifCLASSOPTIONcompsoc
  \section*{Acknowledgments}
\else
  \section*{Acknowledgment}
\fi

This work is jointly sponsored by National Natural Science Foundation of China (U19B2038, 62141608, 62236010, 62206291).


%\bibliographystyle{IEEEtran}

\bibliographystyle{IEEEtran}
\bibliography{IEEEabrv, tkde_2022.bib}

\begin{thebibliography}{10}
\providecommand{\url}[1]{#1}
\csname url@samestyle\endcsname
\providecommand{\newblock}{\relax}
\providecommand{\bibinfo}[2]{#2}
\providecommand{\BIBentrySTDinterwordspacing}{\spaceskip=0pt\relax}
\providecommand{\BIBentryALTinterwordstretchfactor}{4}
\providecommand{\BIBentryALTinterwordspacing}{\spaceskip=\fontdimen2\font plus
\BIBentryALTinterwordstretchfactor\fontdimen3\font minus
  \fontdimen4\font\relax}
\providecommand{\BIBforeignlanguage}[2]{{%
\expandafter\ifx\csname l@#1\endcsname\relax
\typeout{** WARNING: IEEEtran.bst: No hyphenation pattern has been}%
\typeout{** loaded for the language `#1'. Using the pattern for}%
\typeout{** the default language instead.}%
\else
\language=\csname l@#1\endcsname
\fi
#2}}
\providecommand{\BIBdecl}{\relax}
\BIBdecl

\bibitem{xu2022evidenceaware}
W.~Xu, J.~Wu, Q.~Liu, S.~Wu, and L.~Wang, ``Evidence-aware fake news detection
  with graph neural networks,'' in \emph{WWW}, 2022, p. 2501–2510.

\bibitem{Allcott2017SocialMA}
H.~Allcott and M.~Gentzkow, ``Social media and fake news in the 2016
  election,'' \emph{CSN: Politics (Topic)}, 2017.

\bibitem{Naeem2020TheC}
S.~B. Naeem and R.~Bhatti, ``The covid‐19 ‘infodemic’: a new front for
  information professionals,'' \emph{Health Information and Libraries Journal},
  2020.

\bibitem{Sheng2021IntegratingPA}
Q.~Sheng, X.~Zhang, J.~Cao, and L.~Zhong, ``Integrating pattern- and fact-based
  fake news detection via model preference learning,'' \emph{ArXiv}, vol.
  abs/2109.11333, 2021.

\bibitem{popat2018declare}
K.~Popat, S.~Mukherjee, A.~Yates, and G.~Weikum, ``Declare: Debunking fake news
  and false claims using evidence-aware deep learning,'' in \emph{EMNLP}, 2018,
  pp. 22--32.

\bibitem{ma2019sentence}
J.~Ma, W.~Gao, S.~Joty, and K.-F. Wong, ``Sentence-level evidence embedding for
  claim verification with hierarchical attention networks,'' in \emph{ACL},
  2019, pp. 2561--2571.

\bibitem{vo2021hierarchical}
N.~Vo and K.~Lee, ``Hierarchical multi-head attentive network for
  evidence-aware fake news detection,'' in \emph{EACL}, 2021, pp. 965--975.

\bibitem{wu2021unified}
L.~Wu, Y.~Rao, Y.~Lan, L.~Sun, and Z.~Qi, ``Unified dual-view cognitive model
  for interpretable claim verification,'' \emph{ArXiv}, 2021.

\bibitem{Yao2019GraphCN}
L.~Yao, C.~Mao, and Y.~Luo, ``Graph convolutional networks for text
  classification,'' \emph{ArXiv}, vol. abs/1809.05679, 2019.

\bibitem{texting}
Y.~Zhang, X.~Yu, Z.~Cui, S.~Wu, Z.~Wen, and L.~Wang, ``Every document owns its
  structure: Inductive text classification via graph neural networks,''
  \emph{ArXiv}, vol. abs/2004.13826, 2020.

\bibitem{srgnn}
S.~Wu, Y.~Tang, Y.~Zhu, L.~Wang, X.~Xie, and T.~Tan, ``Session-based
  recommendation with graph neural networks,'' in \emph{AAAI}, 2019.

\bibitem{chen2021cil}
T.~Chen, H.~Shi, S.~Tang, Z.~Chen, F.~Wu, and Y.~Zhuang, ``Cil: Contrastive
  instance learning framework for distantly supervised relation extraction,''
  in \emph{ACL}, 2021, pp. 6191--6200.

\bibitem{zhang2021supporting}
D.~Zhang, F.~Nan, X.~Wei, S.-W. Li, H.~Zhu, K.~Mckeown, R.~Nallapati, A.~O.
  Arnold, and B.~Xiang, ``Supporting clustering with contrastive learning,'' in
  \emph{NAACL}, 2021, pp. 5419--5430.

\bibitem{miyato2016adversarial}
T.~Miyato, A.~M. Dai, and I.~Goodfellow, ``Adversarial training methods for
  semi-supervised text classification,'' \emph{ArXiv}, 2016.

\bibitem{li2022hiclre}
D.~Li, T.~Zhang, N.~Hu, C.~Wang, and X.~He, ``Hiclre: A hierarchical
  contrastive learning framework for distantly supervised relation
  extraction,'' in \emph{ACL findings}, 2022, pp. 2567--2578.

\bibitem{Defferrard2016ConvolutionalNN}
M.~Defferrard, X.~Bresson, and P.~Vandergheynst, ``Convolutional neural
  networks on graphs with fast localized spectral filtering,'' in \emph{NIPS},
  2016.

\bibitem{Kipf2017SemiSupervisedCW}
T.~Kipf and M.~Welling, ``Semi-supervised classification with graph
  convolutional networks,'' \emph{ArXiv}, vol. abs/1609.02907, 2017.

\bibitem{Velickovic2018GraphAN}
P.~Velickovic, G.~Cucurull, A.~Casanova, A.~Romero, P.~Lio’, and Y.~Bengio,
  ``Graph attention networks,'' \emph{ArXiv}, vol. abs/1710.10903, 2018.

\bibitem{Hamilton2017InductiveRL}
W.~L. Hamilton, Z.~Ying, and J.~Leskovec, ``Inductive representation learning
  on large graphs,'' in \emph{NIPS}, 2017.

\bibitem{Chen2020HandlingIL}
T.~Chen and R.~C.-W. Wong, ``Handling information loss of graph neural networks
  for session-based recommendation,'' in \emph{KDD}, 2020.

\bibitem{zhang2020personalized}
M.~Zhang, S.~Wu, M.~Gao, X.~Jiang, K.~Xu, and L.~Wang, ``Personalized graph
  neural networks with attention mechanism for session-aware recommendation,''
  \emph{IEEE Transactions on Knowledge and Data Engineering (TKDE)}, 2020.

\bibitem{Zhang2021MiningLS}
J.~Zhang, Y.~Zhu, Q.~Liu, S.~Wu, S.~Wang, and L.~Wang, ``Mining latent
  structures for multimedia recommendation,'' in \emph{ACM Multimedia}, 2021.

\bibitem{Wang2020RelationalGA}
K.~Wang, W.~Shen, Y.~Yang, X.~Quan, and R.~Wang, ``Relational graph attention
  network for aspect-based sentiment analysis,'' in \emph{ACL}, 2020.

\bibitem{Li2021DualGC}
R.~Li, H.~Chen, F.~Feng, Z.~Ma, X.~Wang, and E.~H. Hovy, ``Dual graph
  convolutional networks for aspect-based sentiment analysis,'' in
  \emph{ACL/IJCNLP}, 2021.

\bibitem{Jin2020GraphSL}
W.~Jin, Y.~Ma, X.~Liu, X.~Tang, S.~Wang, and J.~Tang, ``Graph structure
  learning for robust graph neural networks,'' \emph{KDD}, 2020.

\bibitem{zhu2021gsl}
Y.~Zhu, W.~Xu, J.~Zhang, Q.~Liu, S.~Wu, and L.~Wang, ``Deep graph structure
  learning for robust representations: {A} survey,'' \emph{CoRR}, 2021.

\bibitem{jiang2019gsl}
B.~Jiang, Z.~Zhang, D.~Lin, J.~Tang, and B.~Luo, ``Semi-supervised learning
  with graph learning-convolutional networks,'' in \emph{CVPR}, 2019, pp.
  11\,305--11\,312.

\bibitem{chen2020gsl}
Y.~Chen, L.~Wu, and M.~Zaki, ``Iterative deep graph learning for graph neural
  networks: Better and robust node embeddings,'' in \emph{NIPS}, 2020, pp.
  19\,314--19\,326.

\bibitem{cosmo2020gsl}
L.~Cosmo, A.~Kazi, S.~Ahmadi, N.~Navab, and M.~M. Bronstein, ``Latent patient
  network learning for automatic diagnosis,'' 2020.

\bibitem{li2018gsl}
R.~Li, S.~Wang, F.~Zhu, and J.~Huang, ``Adaptive graph convolutional neural
  networks,'' \emph{ArXiv}, vol. abs/1801.03226, 2018.

\bibitem{wu2018gsl}
X.-W. Wu, L.~Zhao, and L.~Akoglu, ``A quest for structure: Jointly learning the
  graph structure and semi-supervised classification,'' \emph{CIKM}, 2018.

\bibitem{franceschi2018gsl}
L.~Franceschi, P.~Frasconi, S.~Salzo, R.~Grazzi, and M.~Pontil, ``Bilevel
  programming for hyperparameter optimization and meta-learning,'' in
  \emph{ICML}, 2018, pp. 1568--1577.

\bibitem{franceschi2019gsl}
L.~Franceschi, M.~Niepert, M.~Pontil, and X.~He, ``Learning discrete structures
  for graph neural networks,'' in \emph{ICML}, 2019, pp. 1972--1982.

\bibitem{Zhang2019gsl}
Y.~Zhang, S.~Pal, M.~J. Coates, and D.~{\"U}stebay, ``Bayesian graph
  convolutional neural networks for semi-supervised classification,'' in
  \emph{AAAI}, 2019.

\bibitem{Yang2019gsl}
L.~Yang, Z.~Kang, X.~Cao, D.~Jin, B.~Yang, and Y.~Guo, ``Topology optimization
  based graph convolutional network,'' in \emph{IJCAI}, 2019.

\bibitem{ying2018hierarchical}
Z.~Ying, J.~You, C.~Morris, X.~Ren, W.~Hamilton, and J.~Leskovec,
  ``Hierarchical graph representation learning with differentiable pooling,''
  in \emph{NIPS}, 2018, pp. 4800--4810.

\bibitem{gao2019graph}
H.~Gao and S.~Ji, ``Graph u-nets,'' in \emph{ICML}, 2019.

\bibitem{lee2019self}
J.~Lee, I.~Lee, and J.~Kang, ``Self-attention graph pooling,'' in \emph{ICML},
  2019, pp. 6661--6670.

\bibitem{Zhou2019GEARGE}
J.~Zhou, X.~Han, C.~Yang, Z.~Liu, L.~Wang, C.~Li, and M.~Sun, ``Gear:
  Graph-based evidence aggregating and reasoning for fact verification,'' in
  \emph{ACL}, 2019.

\bibitem{Liu2020FinegrainedFV}
Z.~Liu, C.~Xiong, M.~Sun, and Z.~Liu, ``Fine-grained fact verification with
  kernel graph attention network,'' in \emph{ACL}, 2020.

\bibitem{Zhong2020ReasoningOS}
W.~Zhong, J.~Xu, D.~Tang, Z.~Xu, N.~Duan, M.~Zhou, J.~Wang, and J.~Yin,
  ``Reasoning over semantic-level graph for fact checking,'' in \emph{ACL},
  2020.

\bibitem{popat2016}
K.~Popat, S.~Mukherjee, J.~Str\"{o}tgen, and G.~Weikum, ``Credibility
  assessment of textual claims on the web,'' in \emph{CIKM}, 2016, p.
  2173–2178.

\bibitem{Yu2017ACA}
F.~Yu, Q.~Liu, S.~Wu, L.~Wang, and T.~Tan, ``A convolutional approach for
  misinformation identification,'' in \emph{IJCAI}, 2017.

\bibitem{volkova-etal-2017-separating}
S.~Volkova, K.~Shaffer, J.~Y. Jang, and N.~Hodas, ``Separating facts from
  fiction: Linguistic models to classify suspicious and trusted news posts on
  {T}witter,'' in \emph{ACL}, 2017, pp. 647--653.

\bibitem{Vo2018}
N.~Vo and K.~Lee, ``The rise of guardians: Fact-checking url recommendation to
  combat fake news,'' in \emph{SIGIR}, 2018, p. 275–284.

\bibitem{Benamira2019SemiSupervisedLA}
A.~Benamira, B.~Devillers, E.~Lesot, A.~Ray, M.~Saadi, and F.~D. Malliaros,
  ``Semi-supervised learning and graph neural networks for fake news
  detection,'' \emph{2019 IEEE/ACM International Conference on Advances in
  Social Networks Analysis and Mining (ASONAM)}, pp. 568--569, 2019.

\bibitem{Chandra2020GraphbasedMO}
S.~Chandra, P.~Mishra, H.~Yannakoudakis, and E.~Shutova, ``Graph-based modeling
  of online communities for fake news detection,'' \emph{ArXiv}, vol.
  abs/2008.06274, 2020.

\bibitem{Jin2021TowardsFR}
Y.~Jin, X.~Wang, R.~Yang, Y.~Sun, W.~Wang, H.~Liao, and X.~Xie, ``Towards
  fine-grained reasoning for fake news detection,'' \emph{ArXiv}, vol.
  abs/2110.15064, 2021.

\bibitem{liu2018mining}
Q.~Liu, F.~Yu, S.~Wu, and L.~Wang, ``Mining significant microblogs for
  misinformation identification: an attention-based approach,'' \emph{ACM
  Transactions on Intelligent Systems and Technology (TIST)}, vol.~9, no.~5,
  pp. 1--20, 2018.

\bibitem{Ajao2019}
O.~Ajao, D.~Bhowmik, and S.~Zargari, ``Sentiment aware fake news detection on
  online social networks,'' in \emph{ICASSP 2019 - 2019 IEEE International
  Conference on Acoustics, Speech and Signal Processing (ICASSP)}, 2019, pp.
  2507--2511.

\bibitem{Gia2019}
A.~Giachanou, P.~Rosso, and F.~Crestani, ``Leveraging emotional signals for
  credibility detection,'' in \emph{SIGIR}, 2019, p. 877–880.

\bibitem{zhang2021www}
X.~Zhang, J.~Cao, X.~Li, Q.~Sheng, L.~Zhong, and K.~Shu, ``Mining dual emotion
  for fake news detection,'' ser. WWW, 2021, p. 3465–3476.

\bibitem{zhu2022memory}
Y.~Zhu, Q.~Sheng, J.~Cao, Q.~Nan, K.~Shu, M.~Wu, J.~Wang, and F.~Zhuang,
  ``Memory-guided multi-view multi-domain fake news detection,'' \emph{IEEE
  Transactions on Knowledge and Data Engineering (TKDE)}, 2022.

\bibitem{vlachos-riedel-2015-identification}
A.~Vlachos and S.~Riedel, ``Identification and verification of simple claims
  about statistical properties,'' in \emph{EMNLP}, 2015, pp. 2596--2601.

\bibitem{vlachos-riedel-2014-fact}
------, ``Fact checking: Task definition and dataset construction,'' in
  \emph{Proceedings of the {ACL} 2014 Workshop on Language Technologies and
  Computational Social Science}, 2014, pp. 18--22.

\bibitem{wu2020evidence}
L.~Wu, Y.~Rao, X.~Yang, W.~Wang, and A.~Nazir, ``Evidence-aware hierarchical
  interactive attention networks for explainable claim verification.'' in
  \emph{IJCAI}, 2020, pp. 1388--1394.

\bibitem{Wu_Rao_Sun_He_2021}
L.~Wu, Y.~Rao, L.~Sun, and W.~He, ``Evidence inference networks for
  interpretable claim verification,'' \emph{AAAI}, pp. 14\,058--14\,066, 2021.

\bibitem{oord2018representation}
A.~v.~d. Oord, Y.~Li, and O.~Vinyals, ``Representation learning with
  contrastive predictive coding,'' \emph{ArXiv}, vol. abs/1807.03748, 2018.

\bibitem{chen2020simple}
T.~Chen, S.~Kornblith, M.~Norouzi, and G.~Hinton, ``A simple framework for
  contrastive learning of visual representations,'' in \emph{ICML}.\hskip 1em
  plus 0.5em minus 0.4em\relax PMLR, 2020, pp. 1597--1607.

\bibitem{he2020momentum}
K.~He, H.~Fan, Y.~Wu, S.~Xie, and R.~Girshick, ``Momentum contrast for
  unsupervised visual representation learning,'' in \emph{CVPR}, 2020, pp.
  9729--9738.

\bibitem{khosla2020supervised}
P.~Khosla, P.~Teterwak, C.~Wang, A.~Sarna, Y.~Tian, P.~Isola, A.~Maschinot,
  C.~Liu, and D.~Krishnan, ``Supervised contrastive learning,'' in \emph{NIPS},
  vol.~33, 2020, pp. 18\,661--18\,673.

\bibitem{bachman2019learning}
P.~Bachman, R.~D. Hjelm, and W.~Buchwalter, ``Learning representations by
  maximizing mutual information across views,'' in \emph{NIPS}, vol.~32, 2019.

\bibitem{velickovic2019deep}
P.~Velickovic, W.~Fedus, W.~L. Hamilton, P.~Li{\`o}, Y.~Bengio, and R.~D.
  Hjelm, ``Deep graph infomax.'' in \emph{ICLR}, vol.~2, no.~3, 2019, p.~4.

\bibitem{you2020graph}
Y.~You, T.~Chen, Y.~Sui, T.~Chen, Z.~Wang, and Y.~Shen, ``Graph contrastive
  learning with augmentations,'' in \emph{NIPS}, vol.~33, 2020, pp. 5812--5823.

\bibitem{zhu2020deep}
Y.~Zhu, Y.~Xu, F.~Yu, Q.~Liu, S.~Wu, and L.~Wang, ``Deep graph contrastive
  representation learning,'' in \emph{GRL+@ICML}, 2020.

\bibitem{zhu2021graph}
------, ``Graph contrastive learning with adaptive augmentation,'' in
  \emph{WWW}, 2021.

\bibitem{gao2021simcse}
T.~Gao, X.~Yao, and D.~Chen, ``Simcse: Simple contrastive learning of sentence
  embeddings,'' in \emph{EMNLP}, 2021, pp. 6894--6910.

\bibitem{lin2022detect}
H.~Lin, J.~Ma, L.~Chen, Z.~Yang, M.~Cheng, and G.~Chen, ``Detect rumors in
  microblog posts for low-resource domains via adversarial contrastive
  learning,'' \emph{ArXiv}, 2022.

\bibitem{yue2022contrastive}
Z.~Yue, H.~Zeng, Z.~Kou, L.~Shang, and D.~Wang, ``Contrastive domain adaptation
  for early misinformation detection: A case study on covid-19,'' \emph{ArXiv},
  2022.

\bibitem{grmm}
Y.~Zhang, J.~Zhang, Z.~Cui, S.~Wu, and L.~Wang, ``A graph-based relevance
  matching model for ad-hoc retrieval,'' in \emph{AAAI}, 2021.

\bibitem{ghrm}
X.~Yu, W.~Xu, Z.~Cui, S.~Wu, and L.~Wang, ``Graph-based hierarchical relevance
  matching signals for ad-hoc retrieval,'' in \emph{WWW}, 2021.

\bibitem{xiong2017end}
C.~Xiong, Z.~Dai, J.~Callan, Z.~Liu, and R.~Power, ``End-to-end neural ad-hoc
  ranking with kernel pooling,'' in \emph{SIGIR}, 2017, pp. 55--64.

\bibitem{Popat2017WhereTT}
K.~Popat, S.~Mukherjee, J.~Str{\"o}tgen, and G.~Weikum, ``Where the truth lies:
  Explaining the credibility of emerging claims on the web and social media,''
  \emph{Proceedings of the 26th International Conference on World Wide Web
  Companion}, 2017.

\bibitem{Rashkin2017TruthOV}
H.~Rashkin, E.~Choi, J.~Y. Jang, S.~Volkova, and Y.~Choi, ``Truth of varying
  shades: Analyzing language in fake news and political fact-checking,'' in
  \emph{EMNLP}, 2017.

\bibitem{lstm}
S.~Hochreiter and J.~Schmidhuber, ``Long short-term memory,'' \emph{Neural
  Computation}, vol.~9, pp. 1735--1780, 1997.

\bibitem{textcnn}
W.~Y. Wang, ``"liar, liar pants on fire": A new benchmark dataset for fake news
  detection,'' in \emph{ACL}, 2017.

\bibitem{bert}
J.~Devlin, M.-W. Chang, K.~Lee, and K.~Toutanova, ``Bert: Pre-training of deep
  bidirectional transformers for language understanding,'' in \emph{NAACL},
  2019.

\bibitem{Li2018DeeperII}
Q.~Li, Z.~Han, and X.-M. Wu, ``Deeper insights into graph convolutional
  networks for semi-supervised learning,'' \emph{ArXiv}, vol. abs/1801.07606,
  2018.

\bibitem{zhang2021latent}
J.~Zhang, Y.~Zhu, Q.~Liu, M.~Zhang, S.~Wu, and L.~Wang, ``Latent structures
  mining with contrastive modality fusion for multimedia recommendation,''
  \emph{Arxiv}, 2021.

\bibitem{yu2021self}
J.~Yu, H.~Yin, J.~Li, Q.~Wang, N.~Q.~V. Hung, and X.~Zhang, ``Self-supervised
  multi-channel hypergraph convolutional network for social recommendation,''
  in \emph{WWW}, 2021, pp. 413--424.

\end{thebibliography}

% H.~Kopka and P.~W. Daly, \emph{A Guide to {\LaTeX}}, 3rd~ed.\hskip 1em plus

% \end{thebibliography}


\begin{IEEEbiography}[{\includegraphics[width=1in,height=1.25in,keepaspectratio]{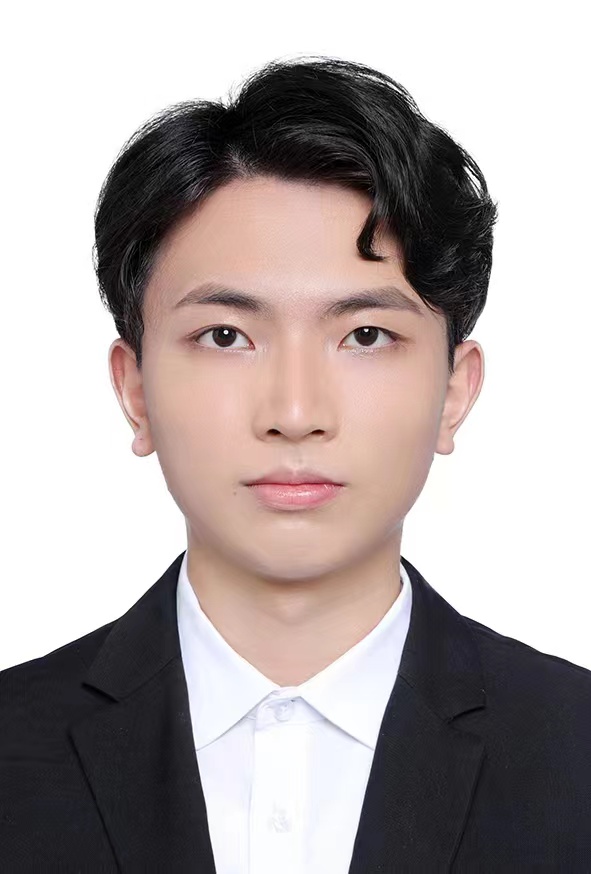}}]{Junfei Wu}
is currently pursuing his Ph.D. degree of Computer Science at the Center for Research on Intelligent Perception and Computing (CRIPAC) at National Laboratory of Pattern Recognition (NLPR), Institute of Automation, Chinese Academy of Sciences (CASIA). His current research interests mainly include fake news detection.
\end{IEEEbiography}

\begin{IEEEbiography}[{\includegraphics[width=1in,height=1.25in,keepaspectratio]{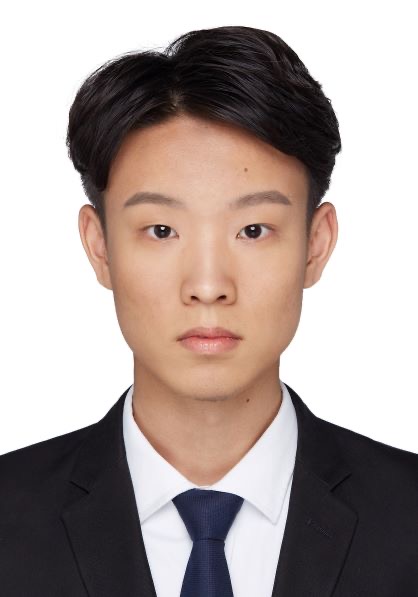}}]{Weizhi Xu}
is currently pursuing his master's degree of Computer Science at the Center for Research on Intelligent Perception and Computing (CRIPAC) at National Laboratory of Pattern Recognition (NLPR), Institute of Automation, Chinese Academy of Sciences (CASIA). His current research interests mainly include graph representation learning, fake news detection and recommender systems.
\end{IEEEbiography}

\begin{IEEEbiography}[{\includegraphics[width=1in,height=1.25in,keepaspectratio]{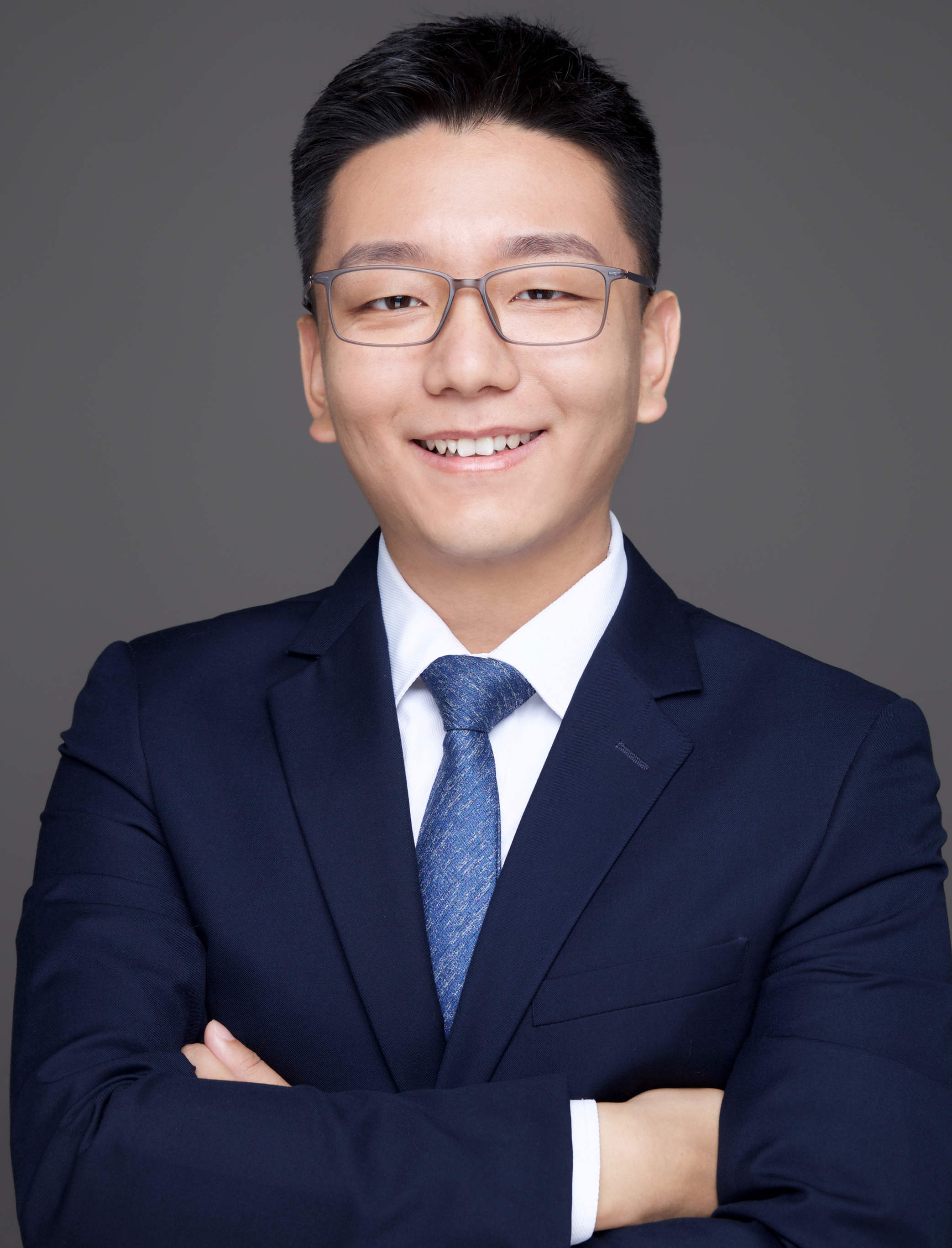}}]{Qiang Liu}
is an Associate Professor with the Center for Research on Intelligent Perception and Computing (CRIPAC), Institute of Automation, Chinese Academy of Sciences (CASIA). He received his PhD degree from CASIA. Currently, his research interests include data mining, recommender systems, text mining, knowledge graph, and graph representation learning. He has published papers in top-tier journals and conferences, such as IEEE TKDE, AAAI, IJCAI, NeurIPS, WWW, SIGIR, CIKM and ICDM.
\end{IEEEbiography}

\begin{IEEEbiography}[{\includegraphics[width=1in,height=1.25in,keepaspectratio]{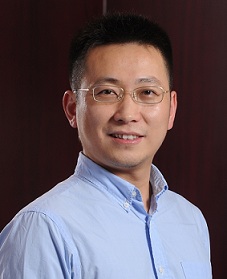}}]{Shu Wu}
received his B.S. degree from Hunan University, China, in 2004, M.S. degree from Xiamen University, China, in 2007, and Ph.D. degree from Department of Computer Science, University of Sherbrooke, Quebec, Canada, all in computer science. He is an Associate Professor with the Center for Research on Intelligent Per- ception and Computing (CRIPAC) at National Laboratory of Pattern Recognition (NLPR), Institute of Automation, Chinese Academy of Sciences (CASIA). He has published more than 50 papers in the areas of data mining and information retrieval in international journals and conferences, such as IEEE TKDE, IEEE THMS, AAAI, ICDM, SIGIR, and CIKM. His research interests include data mining, information retrieval, and recommendation systems.
\end{IEEEbiography}

\begin{IEEEbiography}[{\includegraphics[width=1in,height=1.25in,keepaspectratio]{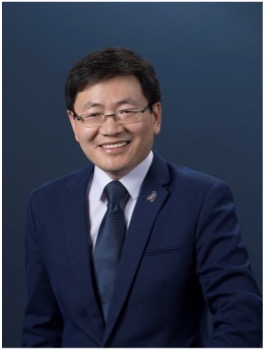}}]{Liang Wang}
received both the BEng and MEng degrees from Anhui University in 1997 and 2000, respectively, and the PhD degree from the Institute of Automation, Chinese Academy of Sci- ences (CASIA) in 2004. From 2004 to 2010, he was a research assistant at Imperial College London, United Kingdom, and Monash University, Australia, a research fellow at the University of Melbourne, Australia, and a lecturer at the University of Bath, United Kingdom, respectively. Currently, he is a full professor of the Hundred
Talents Program at the National Lab of Pattern Recognition, CASIA. His major research interests include machine learning, pattern recog- nition, and computer vision. He has widely published in highly ranked international journals such as IEEE TPAMI and IEEE TIP, and leading international conferences such as CVPR, ICCV, and ECCV. He has served as an Associate Editor of IEEE TPAMI, IEEE TIP, and PR. He is an IEEE Fellow and an IAPR Fellow.
\end{IEEEbiography}


% \newpage


\end{document}